\documentclass[default]{sn-jnl}

\usepackage{graphicx}%
\usepackage{multirow}%
\usepackage{amsmath,amssymb,amsfonts}%
\usepackage{amsthm}%
\usepackage{mathrsfs}%
\usepackage[title]{appendix}%
\usepackage{xcolor}%
\usepackage{textcomp}%
\usepackage{manyfoot}%
\usepackage{booktabs}%
\usepackage{algorithm}%
\usepackage{algorithmicx}%
\usepackage{algpseudocode}%
\usepackage{listings}%
\usepackage{physics}

\DeclareMathOperator*{\argmin}{\arg\min}

\usepackage{geometry}
\theoremstyle{thmstyleone}%
\theoremstyle{thmstyletwo}%

\theoremstyle{thmstylethree}%

\begin{document}

\title[]{Dynamical stability and chaos in artificial neural network trajectories along training}


\author[1]{Kaloyan Danovski\email{kaloyan.danovski@gmail.com}}

\author[1]{Miguel C. Soriano\email{miguel@ifisc.uib-csic.es}}

\author[1]{Lucas Lacasa\email{lucas@ifisc.uib-csic.es}}

\affil[1]{Institute for Cross-Disciplinary Physics and Complex Systems (IFISC, CSIC-UIB), Palma de Mallorca (Spain)}

\abstract{The process of training an artificial neural network involves iteratively adapting its parameters so as to minimize the error of the network's prediction, when confronted with a learning task. This iterative change can be naturally interpreted as a trajectory in network space --a time series of networks-- and thus the training algorithm (e.g. gradient descent optimization of a suitable loss function) can be interpreted as a dynamical system in graph space. 
In order to illustrate this interpretation, here we study the dynamical properties of this process by analyzing through this lens the network trajectories of a shallow neural network, and its evolution through learning a simple classification task. We systematically consider different ranges of the learning rate and explore both the dynamical and orbital stability of the resulting network trajectories, finding hints of regular and chaotic behavior depending on the learning rate regime. Our findings are put in contrast to common wisdom on convergence properties of neural networks and dynamical systems theory. This work also contributes to the cross-fertilization of ideas between dynamical systems theory, network theory and machine learning.}




\maketitle

\section{Introduction}\label{sec1}
The impact of artificial neural network (ANN) models \cite{yegnanarayana2009artificial, goodfellow2016deep} for science, engineering, and technology is undeniable, but their inner workings are notoriously difficult to understand or interpret \cite{marcusDeepLearningCritical2018}. This black-box feeling extends as well to the process of training, whereby an ANN is exposed to a training set to `learn' a representation of the patterns lying inside the input data, with the ultimate goal of leveraging this ANN representation for the prediction of unseen data. In this work we propose to study such a training process --in the scenario where ANNs are used in a supervised learning task-- through the lens of dynamical systems theory, in an attempt to provide mechanistic understanding of the complex behavior emerging in machine learning solutions \cite{san2023frontiers, arola2023effective}. As a matter of fact, training an ANN within a supervised learning task traditionally boils down to an iterative process whereby the parameters of the ANN are sequentially readjusted in order for the output of the ANN to match the expected output of a previously defined ground truth. Such iterative process is, in essence, a (discrete) dynamical system, and more particularly a graph dynamical system \cite{prisner1995graph}, as the mathematical object that evolves through training is the structure of the ANN itself.\\
Note that, from an optimization viewpoint, such dynamics is vastly projected onto a scalar function (the so-called loss function), that training aims to minimize, usually via a gradient-descent type of relaxational dynamics \cite{ruder2016overview}, {i.e. a `pure exploitation' type of search algorithm}. This low-dimensional projection, however, precludes insight on how the specific ANN's structure evolves while the loss function is forced to follow a gradient descent scheme. Therefore, we turn our attention to the question: how does the structure of an ANN evolve in graph space as the loss function is updated?\\
At the same time, observe that not all gradient-descent-like schemes yield necessarily monotonically decreasing loss functions, as this often depends on the particular learning rate within the gradient-descent-based iterative scheme. {More concretely, the learning rate can be adjusted to introduce non-relaxational behavior into the gradient descent dynamics, potentially helping to escape local minima, and thus an element of {\it exploration} is added, making this an exploration-exploitation type of algorithm}.
This raises a second question: How does the specific structure of the ANN evolve in such optimisation schemes that produce non-monotonically decreasing loss functions?

\begin{figure}[htb!]
\centering
	\includegraphics[width=0.7\textwidth]{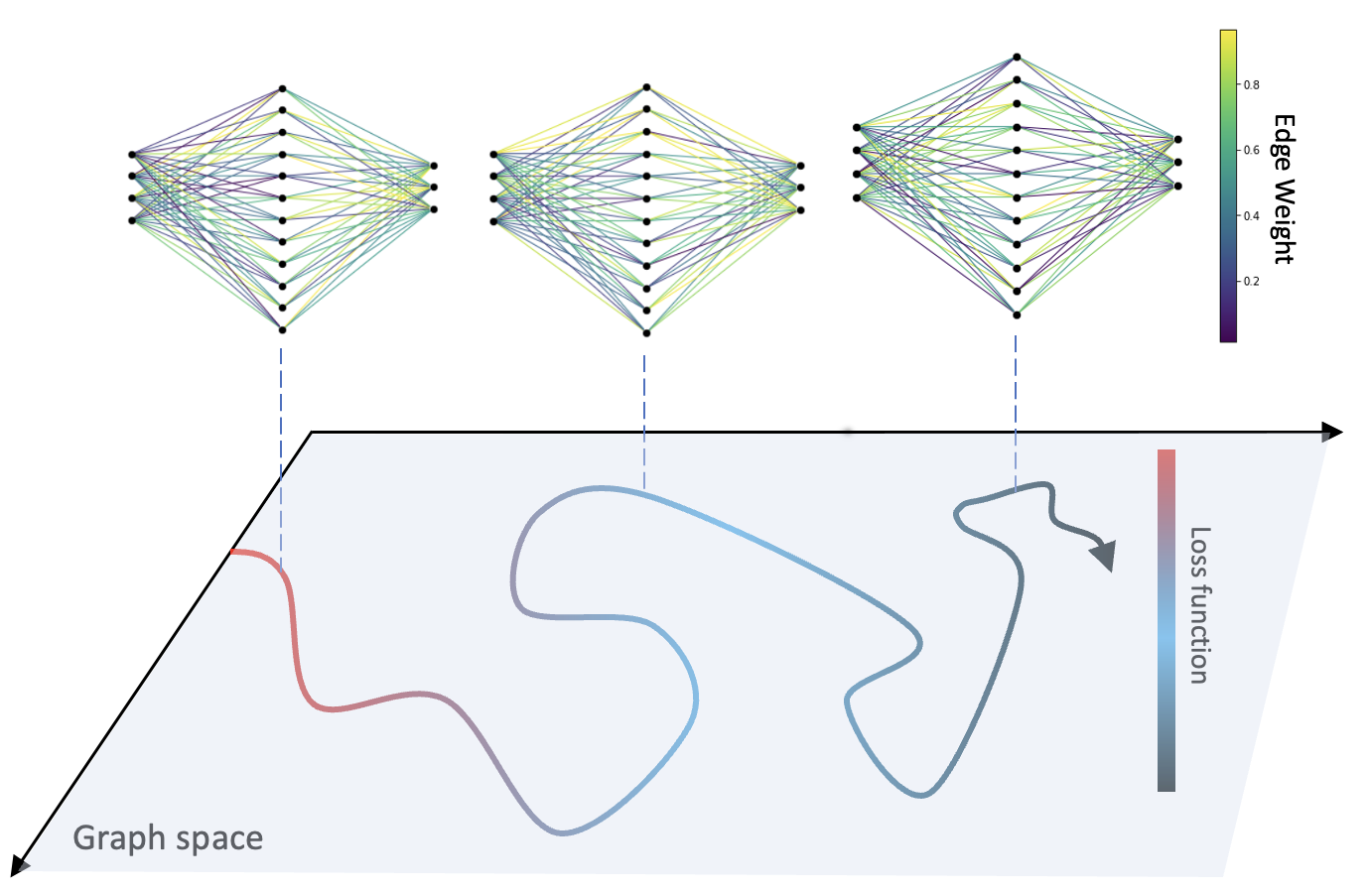}
	\caption{The training process of an ANN is depicted as a network trajectory in graph space, where in each iteration of the optimization scheme the network parameters are updated, leading to a decreasing loss function.}
	\label{fig:cartoon}
\end{figure}

\medskip
\noindent These questions naturally call for the use of dynamical systems concepts and tools, such as dynamical and orbital stability. Under this lens, the ANN is an evolving system whose dynamical variables are the parameters (weights and biases) and the dynamical equations are those implicitly defined by the training algorithm.
The training algorithm is a scheme applied iteratively to batches of input data, in such a way that the ANN parameters are successively updated \cite{hoffer2017train}, i.e. this is a (high dimensional)  discrete-time map. The whole training process is thus nothing but a trajectory in high-dimensional weight space, i.e. a specific type of temporal network \cite{holme2019temporal} that has been coined as a network trajectory \cite{lacasaCorrelationsNetworkTrajectories2022a,caligiuriLyapunovExponentsTemporal2023}, see Fig. \ref{fig:cartoon} for an illustration. The purpose of training is to take the loss function to a minimum which, intuitively, is a stationary point not only of such loss function, but also of the implicitly defined network dynamics. However, it is not clear whether this intuition always holds, or what happens if the training algorithm is designed to produce a non-monotonic evolution of the loss.\\
This paper aims to explore the questions stated above, to challenge some of the basic intuitions, and to further explore the interface between machine learning, dynamical systems and network theory \cite{lamalfaDeepNeuralNetworks2022,lamalfaCharacterizingLearningDynamics2021,ribasFusionComplexNetworks2020,scabiniImprovingDeepNeural2022} by putting together ideas and tools in the specific context of ANN training of a supervised task. The purpose of this work is to offer an illustration of this cross-disciplinary perspective. In particular, we are interested in the dynamical stability of training trajectories in graph space and in the dependence of the dynamical behavior observed on the learning rate parameter of the training algorithm. To this end, inspired by traditional methods and tools from dynamical systems theory \cite{schuster2006deterministic} --such as linear stability theory, the concept of Lyapunov exponents, or orbital stability--, {as well as by more recent ideas of marginal stability close to criticality and the edge of chaos hypothesis \cite{bak1988self, langton1990computation, carroll2020reservoir}},  our main approach is to track, {for specific values of the learning rate}, the evolution (in graph space) of nearby network trajectories during training. This allows us to confirm or dispel some intuitions about the learning process and the landscape of the loss function.

\medskip
\noindent The rest of the paper is organized as follows: in Section \ref{Sec:Preli} we introduce some notation, setting the language and defining the ANN architecture, its specific supervised learning task and its optimization scheme. We discuss how to recast the ANN and its training as a graph dynamical system --establishing trajectories in graph space as orbits-- and, accordingly, introduce relevant tools to investigate its dynamical behavior. In Section \ref{Sec:Preli},  we also discuss related literature in the areas of machine learning, optimization theory as well as dynamical systems. Sections  \ref{Sec:low_learning} and \ref{Sec:high_learning} depict our results on the dynamical stability of the system, which are organized into two separate cases: the low and high learning rate regimes, respectively.
In particular, section \ref{Sec:low_learning} summarises the results of our investigation for the case where the learning rate of the training scheme is sufficiently small so that we expect a monotonically decreasing loss function (purely relaxational dynamics of the loss function). For parsimonious reasons, we keep the ANN topology as simple as possible and propose a feed-forward architecture with a single hidden layer and use a very simple classification task (Iris dataset) as the supervised learning problem. Despite such simplicity, our results challenge basic intuitions of low-dimensional stability theory.
Section \ref{Sec:high_learning} then summarises the results for the opposite case, where the learning rate does not necessarily guarantee convergence of the training algorithm. We find a rich taxonomy of complex dynamical behaviors that non-trivially relate to the actual learning of the ANN. The final part of the manuscript, Section \ref{Sec:discussion}, provides an overview of our findings and relate them to the general questions that we have stated in the opening paragraphs.

\section{Preliminaries}\label{Sec:Preli}

\subsection{Notation, system definition and useful metrics}

To formalize the intuitive notion of ANN training as a graph dynamical system, we first briefly define the network architecture and our training task. A more thorough description can be found in many standard texts on ML and deep learning (e.g. Refs. \cite{yegnanarayana2009artificial,goodfellow2016deep}). As we have mentioned already, in this work we investigate a {quite simple learning task} — a fully-connected, feed-forward neural network with a single hidden layer trained for classification on the Iris dataset \cite{fisher1936use}. A fully-connected, feed-forward artificial neural network (also known as a multi-layer perceptron) is a parametrized functional mapping whose purpose is to encode a meaningful relationship, usually inferred from an empirical dataset during the so-called training process. Conceptually, the non-linear computational units (neurons) of the network can be grouped into layers, where computation flows sequentially through all layers, from input to output.\\
Concretely, we consider a multi-layer ANN as a nonlinear function $F(x; W)$ with input $x$ and parameter set $W$. For a network of $L$ layers (one input layer, one output layer, and $H=L-2$ hidden layers), we define the output at each layer $l=1,...,L$ as $\vb{f}_l \in \mathbb{R}^{n_l}$, where $n_l$ is the dimensionality of layer $l$ (i.e. number of neurons). Thus, $\vb{f}_1=x$ is the input data and $\vb{f}_L=F(x;W)$ is the final output of the network. To obtain the output of layer $l$ for each neuron, we take a weighted sum over the output of the previous layer $l-1$ and feed it through a non-linear activation function $\varphi$. Utilizing vector notation, we have the recursive definition $$\vb{f}_l = \varphi(\vb{W}_l\vb{f}_{l-1}+\vb{b}_l)\qfor l=2,\ldots,L,$$ where $\vb{W}_l$ is an $n_l\times n_{l-1}$ weight matrix and $\vb{b}_l \in \mathbb{R}^{n_l}$ is an additive bias term. In all of our experiments, we use the sigmoid activation function $\varphi(z)=1/(1+e^{-z})$. If the architecture is held fixed, in order to make our functional mapping $F(x;W)$ meaningful, we need to specify the parameter set $W=\{\vb{W}_2,...,\vb{W}_L,\vb{b}_2,...,\vb{b}_L\}$. This is the main problem of training artificial neural networks, often formalized as an optimization task: defining a loss function $\mathcal{L}(x,W)$ that measures the ``badness" of any given output $F(x;W)$ using the parameter set $W$. Then, training is the process that allows you to find 
\begin{align}
\label{eq:opt}
W^* = \argmin_W \mathcal{L}(x,W).
\end{align}
This is usually done via a Gradient Descent optimization algorithm, as described below. In general, the loss function depends on the specific task that we are designing our network to solve. In supervised classification, it quantifies the mismatch between prediction and ground truth, and the standard choice is a cross-entropy loss function $$\mathcal{L}(x,W) = -\frac{1}{N} \sum_i y_i \log F(x_i;W).$$ In general, we can include a so-called regularization term in our loss functions that penalizes large weights in an attempt to prevent overfitting of the model. However, to constrain the complexity of the problem and ease our analysis of the Gradient Descent map, we do not include a regularization term. Later on we discuss the possible implications of this choice.

\begin{figure}
    \centering
    \includegraphics[width=0.6\textwidth]{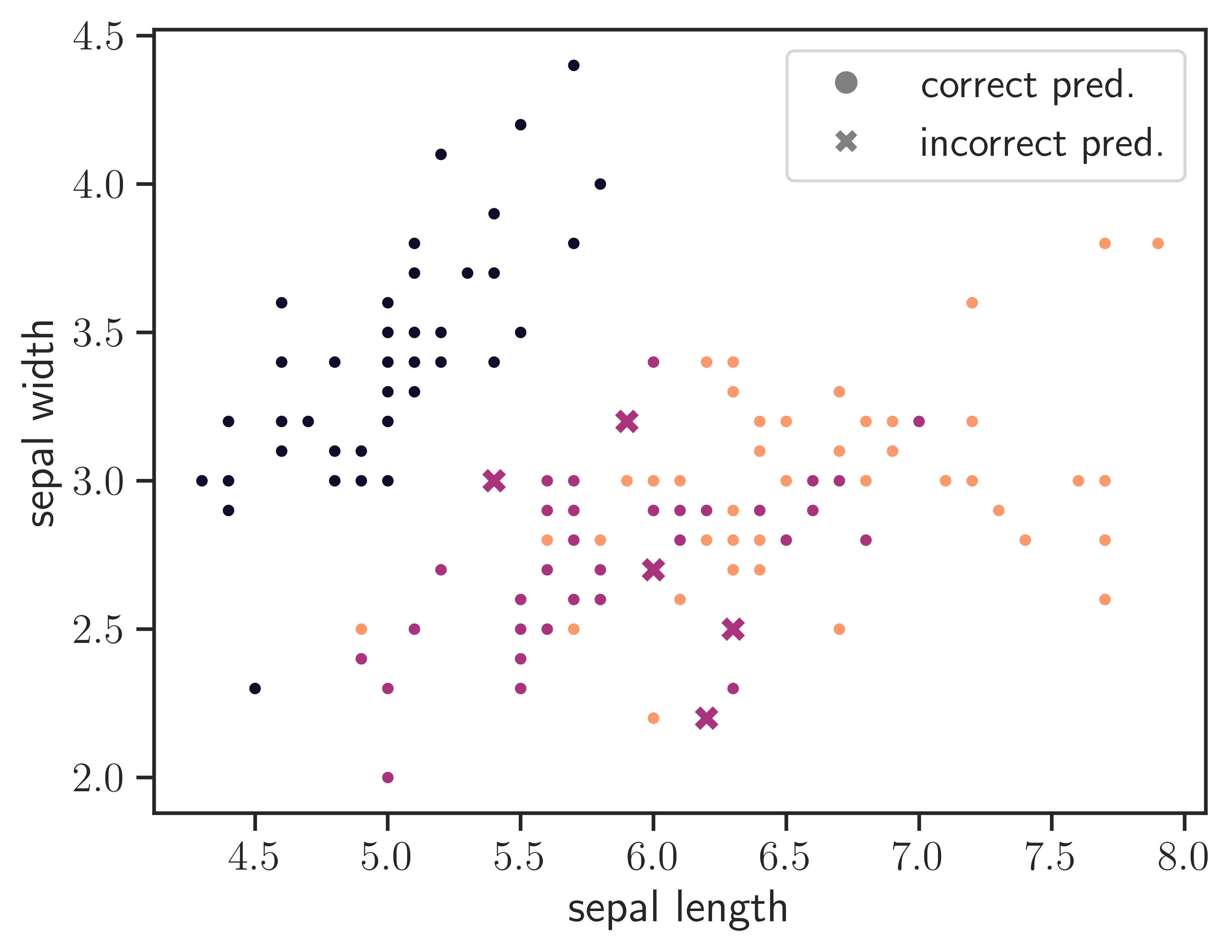}
    \caption{{Illustration of the Iris dataset and difficulty in linearly separating the three classes. Datapoints are shown in the space of two of their four input features, namely ``sepal length" and ``sepal width". Colors correspond to different classes, while markers show whether the instances were classified correctly or not (marked as `x' if the prediction was incorrect).}}
    \label{fig:iris}
\end{figure}

\medskip
\noindent To facilitate the exploration of the training dynamics we have chosen to tackle a simple, toy problem: the so-called Iris dataset (directly available from the scikit-learn library). This dataset consists of $N=150$ samples $\{x_i,y_i\}$, where each $x_i$ is measuring 4 physiological properties of one of three species of \textit{Iris} flowers, and our goal is to train a network that can classify flowers into one of the three species given its physiological properties, i.e. $y_i$ is a categorical variable with three categories. Thus our network has an input dimension $n_1=4$ (number of physiological properties) and an output dimension $n_L=3$. We split the dataset into 120 samples for training and 30 samples for testing performance. {To illustrate why this problem is simple, but non-trivial, in Figure \ref{fig:iris} we plot the dataset in the space of two of its input features, where we can see that some, but not all, of the classes are easily separable. In addition, we mark instances that were not correctly predicted by a typical network trained using the procedure described in this and the following sections, where we can see that these instances fall between two classes whose boundary is ambiguous.}

\medskip
\noindent The choice of a simple task allows us to also use a simple architecture. Our network has a single hidden layer of 10 units. Thus, we have $L=3$, $H=1$, and $n_2=10$. In total, the number of trainable parameters is $\# =83$, which is also the dimensionality of our dynamical system. To initialize our parameters, we set bias weights equal to zero and each of the weights to an independent realization of a standard Gaussian $\hat{N}(0,1)$ random variable.

\medskip
\noindent To solve the optimization problem (\ref{eq:opt}), we employ the Gradient Descent (GD) algorithm, which is used widely in the training of neural networks. Consider any parameter, e.g. $w\in\vb{W}_l$ for any $l$. Under GD, we iteratively update the parameter $w$ based on the gradient of the loss function $\mathcal{L}$ with respect to it, effectively moving $w$ in the direction of the steepest descent along the surface of the loss function. That is, at iteration $t$ we have $$w(t+1)=w(t)-\eta\eval{\pdv{\mathcal{L}}{w}}_{W(t)}.$$ The notation $t$ for the iteration index is not accidental, since we will interpret it as a time parameter when we reformulate the GD algorithm as a discrete dynamical system. We can already see that the above is an equation for a dynamical map. This can be defined equally well for any of the parameter matrices or vectors, where we have $$\vb{W}_l(t+1)=\vb{W}_l(t)-\eta\grad_{\vb{W}_l(t)}\mathcal{L}(x,W(t)).$$ In practice, the gradients are computed using the backpropagation algorithm \cite{goodfellow2016deep}$$\grad_{\vb{W}_l}\mathcal{L}\equiv\pdv{\mathcal{L}}{\vb{W}_l} = \varphi'(\vb{f}_{l-1}) \vb{W}_{l-1} \times \cdots \times \varphi'(\vb{f}_{L-1})\vb{W}_{L-1} \times \varphi'(\vb{f}_L) \vb{W}_L \times \pdv{\mathcal{L}}{\vb{f}_L}.$$

\medskip
\noindent There are in principle many ways to choose the learning rate $\eta$ (e.g. adaptive learning rates \cite{goodfellow2016deep}), but we adopt the simplest strategy, which is to assign $\eta$ a constant value throughout the training process.

\medskip
\noindent In supervised learning, the loss term relies on comparing the outputs of the network $F(x;W)$ given a parameter set $W$ and a set of input data $x$ to the ground-truth of the respective datapoints $x_i \in x$. Given our dataset, how we choose what data to use for calculating the loss term is very important. The most common approach, mainly due to efficiency concerns (see Section \ref{sec:relatedwork}), is to randomly partition the dataset into batches and use the batches for successive parameter updates. This is known as \textit{Stochastic} Gradient Descent (SGD). Alternatively, we can use the whole training dataset at every parameter update step. This is the classic GD algorithm, but in many cases the datasets are too large to make this approach tractable or efficient. However, in our work we use the GD algorithm, which has the advantage of making the dynamics fully deterministic (see \cite{ziyin2023probabilistic} for an interesting analysis on the stability problem for SGD).

\medskip
\noindent 
Finally, we formalize the notion of GD training as a deterministic graph dynamical system that we have alluded to. Since the architecture is very standard, the network is small, and in general here we are more interested in the dynamics of the network throughout training, we shift focus away from the details of the network internals and adopt some notational conveniences. We will represent the parameter set $W$ (i.e. trainable weights) of all $L$ layers as a single composite weight matrix $\vb{W}$. This composite matrix represents the dynamical variables we are interested in, so we define $\vb{W}(t)$ as the weight matrix at time step $t$, i.e. after $t$ iterations of the training algorithm. Thus $\vb{W}(0)$ is the initial condition of our dynamical system, i.e. the (random) initialization of our ANN weights described before. The dynamical equation of our system $\vb{W}(t)$ is the GD algorithm, which now reads:
\begin{align}
\label{eq:gd}
\vb{W}(t+1) = \vb{W}(t) - \eta \grad \mathcal{L}[\vb{W}(t)] \equiv g[\vb{W}(t);\eta].
\end{align}
From this definition, we can immediately form the intuitive conjecture that the stability of the map $\vb{W}(t+1)=g(\vb{W};\eta)$ will depend on the learning rate $\eta$. The rest of our paper is devoted to exploring this dependence, and soon we formalize what we mean by stability.

\medskip

Now that we have formalized our dynamical system, we can define trajectories and a number of tools we will use to study them. Intuitively, the evolution of network weights described by $g(\cdot)$ defines a trajectory through multi-dimensional network space. More formally, the dynamics occur over iterations defined by the composition of the gradient map $g(\vb{W})$, where at time step $t$ the weights are defined as $\vb{W}(t)=g^{(t)}(\vb{W})$ and $g^{(0)}(\vb{W})=\vb{W}(0)$ are the initial weights. We call the sequence $\{\vb{W}(t)\}$ a weight or network \textit{trajectory}, and it is our main object of study. We will sometimes also look at the trajectory of the loss function $\{\mathcal{L}(\vb{W}(t))\}$ and individual weights $\{w_{ij}(t)\}$, all defined intuitively from the network trajectory.\\
Our main approach is to analyze how the distance between initially close trajectories evolves over time. Initially close trajectories are generated by training from an initial condition obtained through a ``small'', point-wise perturbation of a reference weight matrix. Concretely, if our reference matrix is $\vb{W} = \qty{w_{ij}}$, our perturbed matrix will be $\vb{W}'=\{w_{ij} + \delta_{ij}\}$ where the values $\delta_{ij}$ are iid realizations of a random variable $\hat{\delta}$. We can define $\hat{\delta}$ as we wish, but in this work we focus on the case where $\hat{\delta}\equiv\hat{U}(-\epsilon,\epsilon)$. We refer to the parameter $0<\epsilon\ll 1$ as the \textit{perturbation radius}. Intuitively, this perturbation scheme amounts to generating a new set of weights within an $\epsilon$-ball around the reference point.\footnote{To be precise, the perturbation is actually bounded by a hypercube (an $N$-cube) with side length $2\epsilon$ centered at the point defined by $\vb{W}$. Nevertheless, we will still refer to $\epsilon$ as the ``radius'' of perturbations.}\\
Once we obtain a perturbed matrix $\vb{W}'$, we independently train the model with those weights as initialization, to obtain a perturbed trajectory $\{\vb{W}'(t)\}$. Given a reference and perturbed trajectories, we can measure the divergence between them by applying a distance metric $d(\vb{W}(t),\vb{W}'(t))$ at each iteration $t$. We take $d$ as simply the $L_1$-norm of the element-wise difference between weight matrices:
\begin{align}
	d(\vb{W},\vb{W}') = \|\vb{W}-\vb{W}'\| = \sum_{i,j} |w_{ij}-w'_{ij}|.
\end{align}
In a slight abuse of notation, when it is unambiguous to do so we will use $d(t) \equiv d(\vb{W}(t),\vb{W}'(t))$ to denote the evolution of the distance between two trajectories.

\medskip
\noindent In this work we consider two types of stability: stability of stationary solutions (dynamical stability) and stability of network trajectories (orbital stability). 
Dynamical stability refers to understand how $d(t)$ grows or shrinks when $\{\vb{W}\}$ is a network configuration found after the training has finished (i.e. $\{\vb{W}\}$ is somewhat stationary) and $\{\vb{W}'\}$ is a perturbation around the stationary solution.
Orbital stability on the other hand does not require  $\{\vb{W}\}$ to be a stationary solution, and assesses if a perturbed trajectory $\{\vb{W}'\}$ does not diverge (indefinitely) from the reference trajectory $\{\vb{W}\}$: in that case the latter is stable. Of course, this could imply a weaker form of stability, where the perturbation remains bounded within some neighborhood of the reference, or a stronger one if the perturbation is attracted to the reference, i.e. their distance obeys $\lim_{t\to\infty}d(t)=0$.
If the distance between trajectories increases (e.g. exponentially), the reference is dynamically unstable. This kind of sensitive dependence to initial conditions is a feature of many chaotic systems \cite{strogatzNonlinearDynamicsChaos2015}. To study this formally, we will estimate the (network) maximum Lyapunov exponent \cite{caligiuriLyapunovExponentsTemporal2023}, which provides an average estimation of the exponential expansion of initially close network trajectories.
More concretely, given an initial condition (reference trajectory ${\bf W}$) and an ensemble of $M$ perturbed initial conditions, we first calculate the distance $d_i(t), i=1,...,M$ between the reference and each of the $M$ perturbations at each iteration $t$. We then calculate the expansion rate of the distance averaged over perturbations as 
\begin{equation}
  \Lambda({\bf W}) = \frac{1}{\tau}\ln\frac{M^{-1}\sum_{j=1}^M d_j(\tau)}{M^{-1}\sum_{j=1}^M d_j(0)}, \label{eq:Lambda} 
\end{equation}
where the parameter $\tau$ is the saturation time \cite{caligiuriLyapunovExponentsTemporal2023}, at which the distance reaches the size of the attractor and any eventual  exponential divergence necessarily stops. The quantity $\Lambda$ characterizes the local expansion rate around the initial condition ${\bf W}$, and is called a \textit{finite} network Lyapunov exponent {(just as it happens for finite time Lyapunov exponents \cite{aurell1996growth,aurell1997predictability}, here the initial distance between nearby networks cannot --by construction-- be infinitesimally small, and thus the distance between initially close conditions reach the size of the attractor in finite time)}. To obtain a global picture we can look at the distribution $P(\Lambda)$ over different initial conditions. If $P(\Lambda)$ is unimodal, its mean provides an estimate of the \textit{maximum} network Lyapunov exponent $\lambda_{\text{nMLE}}$ \cite{caligiuriLyapunovExponentsTemporal2023}
$$\lambda_{\text{nMLE}} =\langle \Lambda({\bf W})\rangle_{{\bf W}}.$$
The saturation time $\tau$ is fixed for a given reference trajectory and set of perturbations. In practice, it has to be found by visually exploring the distance curve or (better) by numerically estimating a good window for calculating the expansion rate (i.e. by trying various windows and taking the ones with the best exponential fit).


\subsection{Relevant concepts}\label{sec:relatedwork}

\noindent {\bf On convergence of Gradient Descent -- } Since we adopt the perspective of dynamical stability, a particularly relevant question for us is under what conditions GD will converge to a local or global minimum of the loss function, which we will identify with a stable equilibrium of the dynamics. 
Convergence to global minima is guaranteed under the mathematical assumptions of $\ell$-smoothness and (strong) convexity of the loss function $\mathcal{L}$. In those cases GD is guaranteed to make progress towards the global minimum, and if the loss function is convex the convergence is exponential, as we would expect for a globally attracting fixed point. In a sense then the loss function is a Lyapunov function of the dynamical system, as its value will decrease monotonically on trajectories. This is an interesting thought experiment, but this scenario is not realistic. In practically all applications of neural networks, the loss function $\mathcal{L}$ is highly non-convex, with a large number of local minima and saddle points, which represent critical points of the dynamics. {Gradient descent (let it be full-batch, or stochastic) is in general only guaranteed to converge to a local minimum in this case. Interestingly, escape from local minima can be done by accurately striking a balance between so-called {\it exploitation strategies} (e.g. purely relaxational, gradient descent) and {\it exploration rules}. While classic GD with small learning rate can be seen as a pure exploitation strategy, to some extent some variations of such classical scheme such as increasing the learning rate, adding a dropout mechanism, or moving from full-batch GD to stochastic gradient descent, can themselves be seen as adding a certain amount of exploration. Other approaches that fully balance exploration and exploitation include simulated annealing or genetic algorithms \cite{montana1989training}. While finding the global minimum in a highly non-convex loss landscape is generally intractable, in practice this is not an issue of practical concern since for large networks most local minima have near-optimal loss \cite{choromanskaLossSurfacesMultilayer2015}, and therefore exploration rules are not as important as exploitation strategies when it comes to training ANNs}.\\
Convergence theorems give a bound to the learning rate $\eta\leq2/\ell$ (where $\ell$ refers to the Lipschitz constant of the loss function), above which GD is expected to diverge. This bound is commonly used as a heuristic for the choice of learning rate $\eta$, but in practice, a number of studies have observed that learning and convergence towards minima can happen for large $\eta$ at or above this threshold as well \cite{cohenGradientDescentNeural2021,kongStochasticityDeterministicGradient2020,agarwalAccelerationFractalLearning2021}. Exploring this question is one of the goals of this work.\\
An important and often debated question in the literature on optimization theory is the convergence to saddle points.
Theoretically, if saddle points are strict (i.e. at least one of the eigenvalues of the loss function Hessian is strictly negative) then GD will never remain trapped in them \cite{leeGradientDescentOnly2016}, though it is still possible that in practice the trajectories take an impractically long time to escape. Note that for shallow networks with one hidden layer (which we consider in this work) saddle points are guaranteed to be strict \cite{kawaguchiDeepLearningPoor2016,zhuGlobalOptimizationGeometry2020}. However, for the loss function arising in an arbitrary ANN problem it seems difficult to determine whether saddles can be assumed strict or not, so even the convergence of GD towards local minima is not guaranteed theoretically. This is why in practice one never waits for the algorithm to converge to a truly stationary point, but stops training when the gradient has been deemed sufficiently small.

\medskip
\noindent {\bf Loss landscape --} There is also a line of study in machine learning theory focusing on describing the landscape of the loss function, which can give us some insights. In general, studies continuously find that loss landscapes are very difficult to characterize in theory, but seem to behave more simply in practice, and one of our aims is to see if the dynamical systems perspective can help explain this.
In particular, the modality of the loss function (i.e. presence and nature of fixed points) has been studied by many authors. For example, Kawaguchi \cite{kawaguchiDeepLearningPoor2016} shows that all local minima have the same loss and deep networks can have non-strict saddle nodes, which could help to explain why obtaining ``good" results under gradient optimization is tractable despite it being an NP-complete problem in theory. In other work, Bosman et al. \cite{bosmanVisualisingBasinsAttraction2020} among many other authors show that with an increased dimensionality of the ANN problem, the loss landscape contains more saddles and fewer local minima. Lastly, other features of the loss surface seem simpler than we might assume, given that for example regions with low loss are represented by high-dimensional basins rather than isolated points \cite{fortLargeScaleStructure2019}, SGD usually quickly takes solutions to those basins, and then slowly moves to find the most optimal solution \cite{fortDeepEnsemblesLoss2019,havasiTrainingIndependentSubnetworks2020}, and quadratic approximations to the loss landscape do well in the latter phase \cite{fortDeepLearningKernel2020}.

\medskip
\noindent {\bf Reminder on dynamical and orbital stability --} See \cite{strogatzNonlinearDynamicsChaos2015,alligoodChaosIntroductionDynamical1996,Holmes:2006} for background. We restrict our focus to autonomous, discrete-time maps, as is the GD algorithm defined in Eq. \ref{eq:gd}.
Consider a dynamical system in $m$ dimensions with dynamical variable $\vb{x}=(x_1,\dots,x_m)^T \in \mathbb{R}^m$ and the set of difference equations $\vb{f}(\vb{x})$ such that $$\vb{x} \mapsto \vb{f}(\vb{x})\qor \vb{x}_n=\vb{f}(\vb{x}_{n-1})=\vb{f}^n(\vb{x}_0),$$ where $n$ is used to index time as iterations of the map functions $\vb{f}(\vb{x}) = (f_1(\vb{x}),\dots,f_m(\vb{x}))^T$. A fixed point of the system is $\vb{x}^*$ for which $\vb{x}^* = \vb{f}(\vb{x}^*)$. A fixed point is considered \textit{Lyapunov stable} if, intuitively, all orbits starting near the fixed point remain close to it indefinitely, i.e. the distance between points on the trajectory and the fixed point is bounded. Concretely, if for every neighbourhood $U$ around $\vb{x}^*$ there exists a neighbourhood $V \subseteq U$ s.t. $\forall \vb{x}_0 \in V$ we have $\vb{f}^n(\vb{x}_0) \in U$ as $n \to \infty$, the point $\vb{x}^*$ is Lyapunov stable. A point is considered \textit{asymptotically stable} if it is both Lyapunov stable and also $\forall \vb{x} \in V$, $\lim_{n\to\infty} |\vb{f}^n(\vb{x}_0) - \vb{x}^*| = 0$. That is, the distance between nearby points and the fixed point decreases over time, thus $\vb{x}^*$ is attracting. Note this implies that $\frac{|\vb{f}(\vb{x})-\vb{x}^*|}{|\vb{x}-\vb{x}^*|} < a$ for some $0<a<1$, and for a sequence of $n$ iterations of the map we have $|\vb{f}^n(\vb{x})-\vb{x}^*|\leq a^n|\vb{x}-\vb{x}^*|$, which is an exponential convergence to the fixed point $\vb{x}^*$. If a point is Lyapunov stable, but not asymptotically stable, we will call it marginally or neutrally stable, since nearby trajectories neither grow unboundedly nor decrease exponentially.\\
Finally, the concept of \textit{orbital stability} \cite{Holmes:2006} extends the notion of stability of fixed points to general orbits of a dynamical system. The intuition is the same — an asymptotically stable orbit will attract nearby orbits, i.e. the distance between the orbits will shrink over time; while a marginally stable orbit will have nearby trajectories bounded within some neighborhood. This is the main perspective we will adopt when studying the trajectories defined by the GD map in network space.

\section{The low learning rate regime}\label{Sec:low_learning}

Unless otherwise stated, the data we show in this section is based on simulations of (deterministic) Gradient Descent (i.e., full batch) with a small learning rate $\eta=0.01$, a regime in which the convergence of the loss function is typically guaranteed (note however that the precise values of $\eta$ that delineate the different regimes likely depend on the problem, architecture, and data, so the values here should not be taken as universal). \\
{Note that in this work we are more interested in establishing a methodology to study the evolution of network trajectories, rather than in designing optimal ANN architectures. Nevertheless, we recognize that examining the performance of the network on both training and held-out testing data is an important indicator to ensure its architecture is relevant, and thus include sample performance metrics in Appendix \ref{app:iris}.}

\subsection{Divergence of network trajectories and orbital stability}

\begin{figure}[htb!]
	\includegraphics[width=\textwidth]{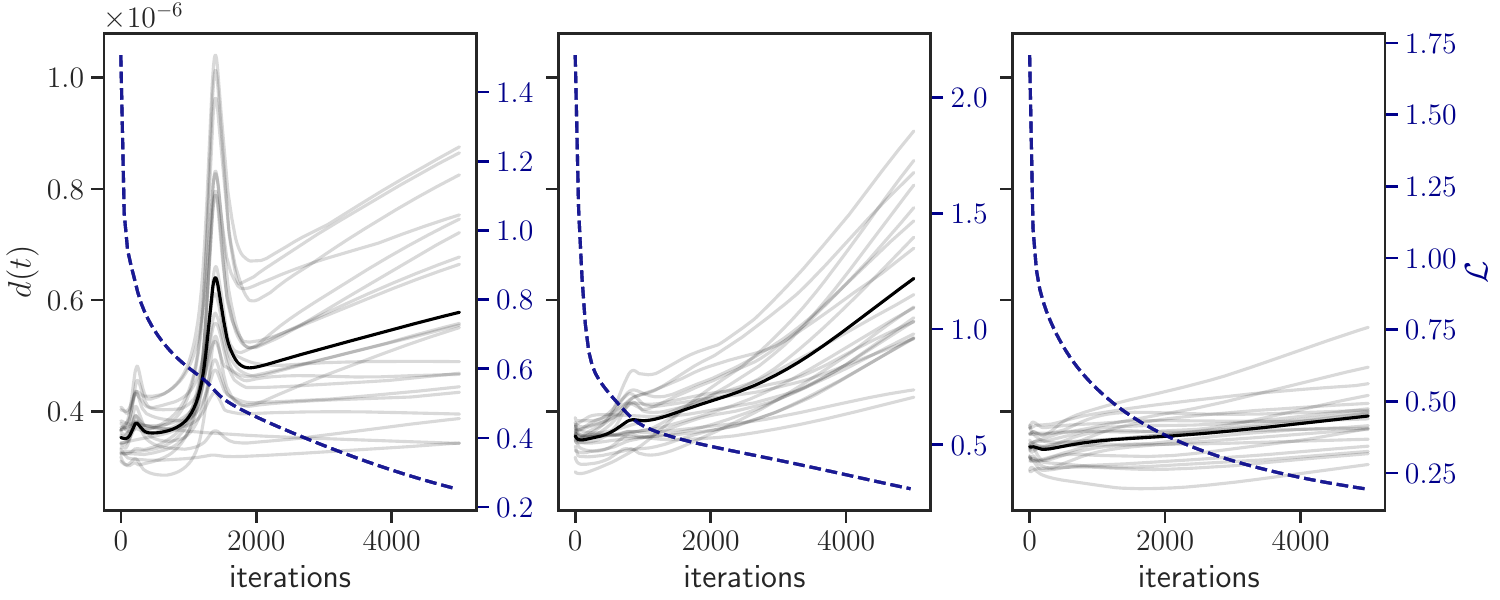}
	\caption{Example showing the evolution of the distances between reference and perturbed trajectories, for a perturbation radius $\epsilon=10^{-8}$. Each panel shows results for a different network initial condition and a random set of perturbations. Gray lines are the distances from individual perturbations, and black is the average distance over all (20) perturbations. Overlaid in blue dashed line (right-hand axis) is the loss trajectory of the network plotted for all perturbations (all loss curves coincide).}
	\label{fig:ex}
\end{figure}

Here, we track and examine the trajectories of the ANN as it learns. For illustration, Fig. \ref{fig:ex} shows the distance $d(t)$ of perturbed network initial conditions with respect to a reference initial condition, the average distance of the ensemble, and training loss $\mathcal{L}$, over the network trajectory depicted through learning (i.e. at each iteration), for a perturbation radius $\epsilon=10^{-8}$. 
While the network is always `learning' (the loss function monotonically decreases), and contrary to naive expectations, we observe that:
\begin{itemize}
    \item The network distance $d(t)$ is not monotonic, and neither increases exponentially (as in chaotic systems), nor monotonically shrinks over time.
    \item For a particular network initial condition, the dynamical evolution of nearby perturbations (within the perturbation radius) is not consistent. This a priori suggests lack of orbital stability. 
    \item The shape of the network distances is dependent on the specific network initial condition, and the dynamical behavior is therefore not ergodic.
\end{itemize}


\begin{figure}[htb!]
    \centering
    \makebox[\textwidth]{\includegraphics[width=1.1\textwidth]{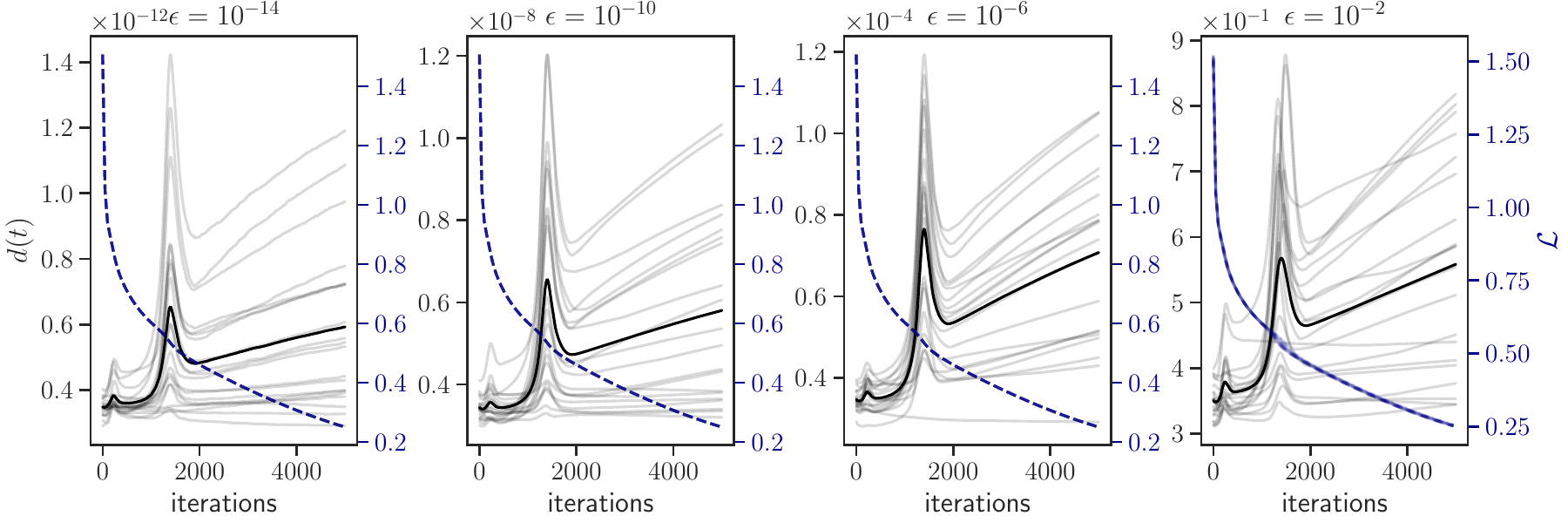}}
    \caption{Evolution of distances between perturbations and reference trajectory for a single initial condition and different values of the perturbation range $\epsilon=\{10^{-14},10^{-10},10^{-6},10^{-2}\}$. Random perturbations are sampled separately for each value of $\epsilon$. Gray lines represent individual perturbations and black line is the mean over perturbations. Note the different scales on the distance axis (left-hand side). The loss of perturbations is overlaid in dashed blue line (right-hand axis).}
	\label{fig:eps_diff}
\end{figure}

A possible explanation for the fact that distances do not vanish (actually, they seem to systematically increase in the long run) and that perturbations exhibit different convergence patterns relative to the reference trajectory would be that, within the perturbation radius, different perturbations are indeed converging towards different minima of the loss function, i.e. different network configurations with nearly-identical loss values \cite{choromanskaLossSurfacesMultilayer2015}.
If this was the main reason underlying the observed phenomenology, then we speculate that, for small enough perturbation radius $\epsilon$, we should find a transition to non-increasing distances. However, the results shown in Fig. \ref{fig:eps_diff} indicate that, for the same network initial condition, perturbations within systematically smaller radius $\epsilon$ still show a similar qualitative behavior for $d(t)$, i.e. almost independently of $\epsilon$. In Fig. \ref{fig:eps_diff}, we do not see a transition between convergence and divergence with $\epsilon$, at least for the values considered here. The conclusion is thus that, numerically, what we are observing is consistent with the lack of orbital stability: there is no small enough $\epsilon$ such that orbits within that radius stay confined.
At the same time, the evolution of network distances is not consistent with sensitive dependence of initial conditions (exponential expansion): enforcing a low learning rate $\eta$ guarantees convergence of the iteration scheme and thus such exponential expansion is not to be expected in this regime.

\begin{figure}[htb!]
	\centering
	\includegraphics[width=0.5\textwidth]{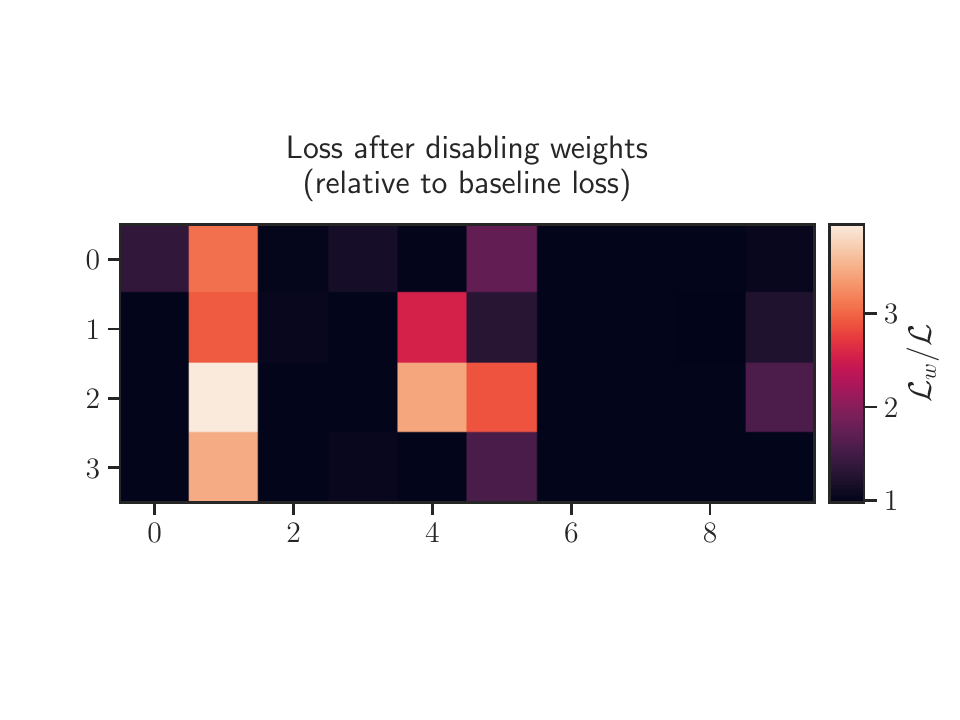}
 	\includegraphics[width=0.49\textwidth]{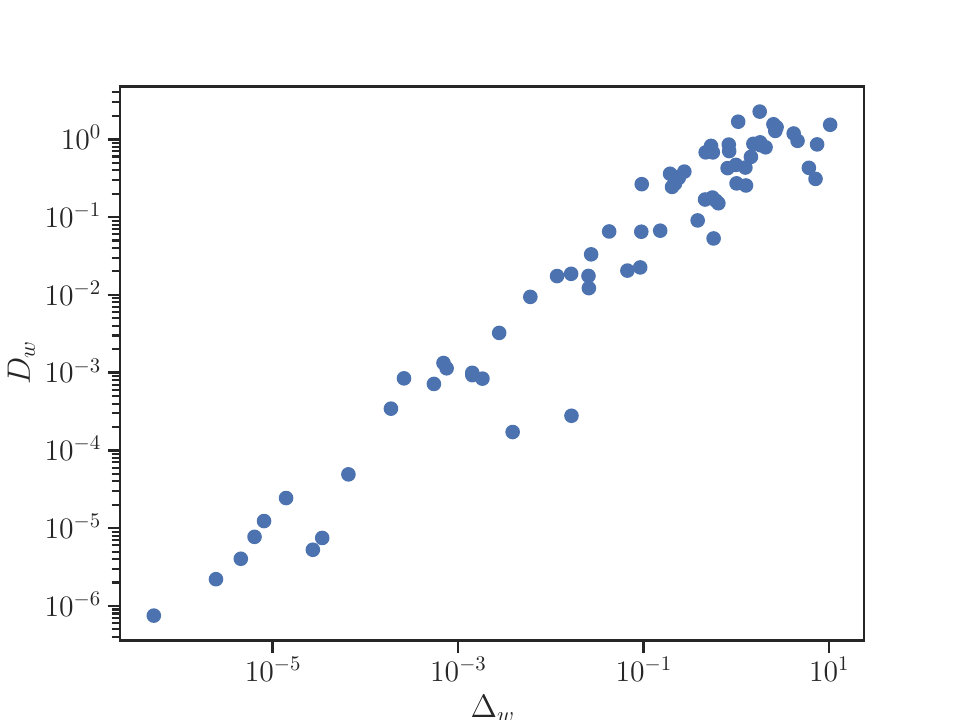}
	\caption{(Left panel) Heatmap for one of the weight matrices of a final solution, i.e. $\bf{W}_1$ at the final iteration. Color corresponds to difference in loss $\mathcal{L}_w$ after disabling each weight individually (i.e. setting $w_{ji}=0$ only and recalculating loss). Difference is shown relative to baseline loss $\mathcal{L}$ when all weights are kept as is. (Right panel) Relationship between per-weight displacements $\Delta_w$ and the total distance travelled by individual weights $D_w$. Results are for the weight matrices of a single network trajectory.}
	\label{fig:w_loss}
\end{figure}

\medskip
\noindent Hence, why do initially close network trajectories appear to continuously diverge, irrespective of the initial distance $\epsilon$? What is causing this lack of orbital stability? This brings us to the conjecture of the existence of irrelevant directions (i.e. that not all weights are important for the output of the network) and the scenario of marginal stability, where the network trajectory would be ``drifting'' along some dimensions that are essentially flat with respect to the loss function, which in turn would imply that minima of the loss function would be represented by stationary manifolds, rather than simple stationary points. In such a scenario, throughout training initially a handful of network parameters would be updating to go in the direction of strong gradients, and eventually most of the gradients would be very small, so that the update would be producing a sort of random walk trajectory within the stationary manifold, and such network diffusion would in turn produce the continuous --yet slow-- divergence of trajectories observed in Fig. \ref{fig:ex} and Fig. \ref{fig:eps_diff}.\\
The existence of irrelevant directions seems to be supported when looking at how the loss changes when we disable individual weights after training has finished. The results in the left panel of Fig. \ref{fig:w_loss} show that the ``importance'' of weights, defined as the decrease in loss incurred after they have been disabled (set to zero), is not normally distributed, and while disabling some weights has a disproportionately large impact on the loss, this impact is minimal for plenty of others. Note, however, that the story is far more intricate, since the ``problem-solving ability'' of neural networks is based not on individual weights, but on non-linear combinations of multiple weights (this echoes the issues with similar ``feature importance'' approaches in classical machine learning).

\noindent To test for the neutral drift hypothesis, we finally look at the dynamics on the level of individual network weights: for each weight $w$, we compute its per-weight displacement $\Delta_w$ as $$\Delta_w=\frac{|w(T)-w(0)|}{|w(0)|},$$ where $T$ denotes the time of the final iteration, and scatter plot it against the per-weight total distance travelled $D_w$, i.e. a rectification of its trajectory $$D_w = \sum_{t=1}^T |w(t)-w(t-1)|.$$ 
Intuitively, along the irrelevant dimensions (with unbiased random-walk-like drift), we would expect a small displacement $\Delta_w$ and a large distance traveled $D_w$, whereas more ballistic weight trajectories would yield $D_w \propto \Delta_w$. The right panel of Fig. \ref{fig:w_loss} shows scatter plots of this relationship for all the weights of a typical network trajectory, finding a linear relationship in this representation between displacement and distance and the lack of weights with both small displacement and large distance. Thus, naive neutral drift is an unlikely explanation.

\medskip
\noindent In the end, the most plausible explanation of the continued divergence observed at perturbation near the initial conditions might come in light of findings about the effect of cross-entropy loss on the shape of low-loss basins \cite{fortDeepEnsemblesLoss2019}. Essentially, once a training algorithm has found a basin of low loss (i.e. classifies samples correctly), it will further try to scale up the outputs of neural units in order to increase the gap with the incorrect prediction and make the outputs match as best as possible the ground-truth (represented by unit vectors with a value of 1 at the index of the correct prediction). As a result of this, low-loss basins for cross-entropy loss ``extend outwards'' from the origin and the dynamics move gradually (but slowly) away from the origin. Since independent initial conditions are unlikely to converge towards the same minima, the perturbations drift away from one another.

\subsection{Stability analysis near the stationary state (post-learning) }
\label{ssec:post}

\begin{figure}[]
	\includegraphics[width=\textwidth]{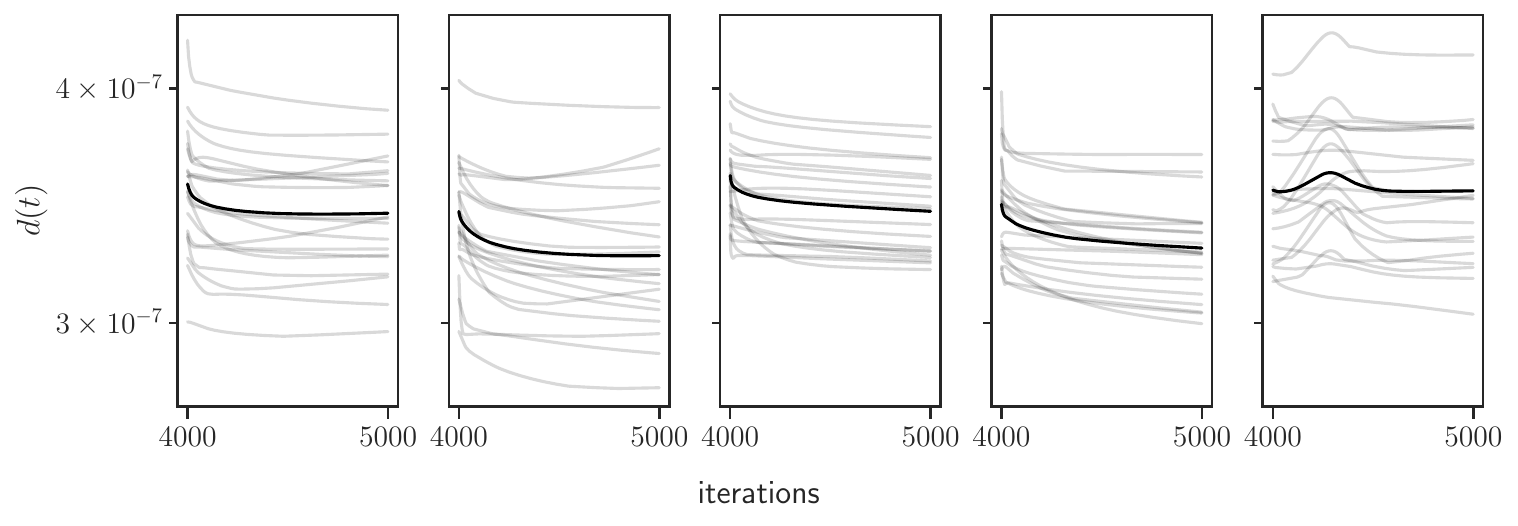}
	\caption{Distance between perturbed and reference trajectories for $\epsilon=10^{-8}$, where perturbations are taken after 4000 learning iterations (the change in the loss function is $O(10^{-2})$). Gray lines correspond to individual perturbations and black line corresponds to mean over perturbations. Each panel shows a different independent initial condition.}
	\label{fig:post}
\end{figure}

In the preceding section we have analysed how perturbations of an initial network condition evolve through learning, i.e. exploring network divergences related to the ideas of orbital stability. Here we turn our attention to the problem of dynamical stability close to stationary points of the dynamics.
It has been observed empirically that at the start of training the network is looking for a basin of lower loss, while later on it is exploring within that region \cite{fortDeepEnsemblesLoss2019}. We would like to explore here the intuition that as the network asymptotically reaches a plateau of the loss function late in training, the trajectory reaches a (stable) stationary solution. 
A naive linear stability theory tells us that, close to a stable (unstable) fixed point, small perturbations generate orbits that converge exponentially towards (diverge exponentially away from) the fixed point, depending on the eigenvalues of the linearized system's Jacobian (the same phenomenology is expected for gradient descent under a strongly convex function).
In Fig. \ref{fig:post} we show examples for the distance $d(t)$ between perturbations taken with respect to a network condition found near the end of training (after 4000 epochs). We see that distance between trajectories predominantly remains flat or decreases slowly, but on rare occasions also increases or experiences bumps. By plotting the distances on a semi-logarithmic scale, we see that for many initial conditions they exhibit a relatively sharp drop followed by a very slow decay. 
These results seem to be incompatible with an exponential shrinkage, and thus the solutions obtained after 4000 training epochs are not strictly attractive fixed points in the sense of dynamical systems, or minima of locally convex functions, from the point of view of optimisation theory.

\noindent The above notwidthstanding, the behavior of the distances shown in Fig. \ref{fig:post} is indeed very different from that observed for perturbations taken at the initial condition, as shown earlier in e.g. Fig. \ref{fig:ex}. Considering the tendency for distances to experience abrupt swelling and other non-trivial behaviors, we cannot state with certainty whether trajectories are bounded for long times, and therefore stable (in a weaker, Lyapunov sense). However, we should be more confident in this conclusion than for perturbations at the initial condition. 

The fact that for the same initial condition some perturbations decay while others grow seem to imply that, within a radius $\epsilon$, the behavior is not homogeneous
(it is still unclear whether perturbed trajectories converge towards the same local minimum).
When minima are locally convex/well-like, the loss function represents a Lyapunov potential and linear stability theory predicts exponential decay, as mentioned before. In the case when minima are not points dotting the phase space, but instead are represented by high-dimensional manifolds, random perturbations would converge to points at a distance from the reference along the flat dimensions. In this case, the distance between some weight values would decay exponentially while for others it would remain constant. In the language of linear stability theory of maps, it is possible that the Jacobian of the linearized system has many eigenvalues equal to 1 and only a handful smaller than 1.\\ 
Finally, since distances do not decay fully, this could again imply that trajectories converge to nearby saddle points. We know that saddle points are ubiquitous in high-dimensional systems (both general dynamical systems \cite{fyodorovNonlinearAnalogueMay2016,benarousCountingEquilibriaLarge2021} and empirically in neural network loss landscapes \cite{bosmanLossSurfaceModality2020}) and that the convergence of GD slows down near any critical point, whether that is a minima or a saddle. In fact, GD is notoriously bad at escaping saddle points (one of the reasons SGD is preferred in practice). On one hand, perturbations away from a local minimum (if that minimum is isolated and narrow) could cause the trajectory to evolve towards nearby saddle-points and slow down the dynamics. On the other hand, perturbations away from saddle points might allow a trajectory to escape that saddle more quickly and continue the evolution towards a different critical point 
(unlikely for low-rank saddles). We return to this discussion in Section \ref{Sec:discussion}, after presenting the exploration of large learning rates in the following section.

\begin{figure}[p]
    \centering
    \makebox[\textwidth]{\includegraphics[width=1.2\textwidth]{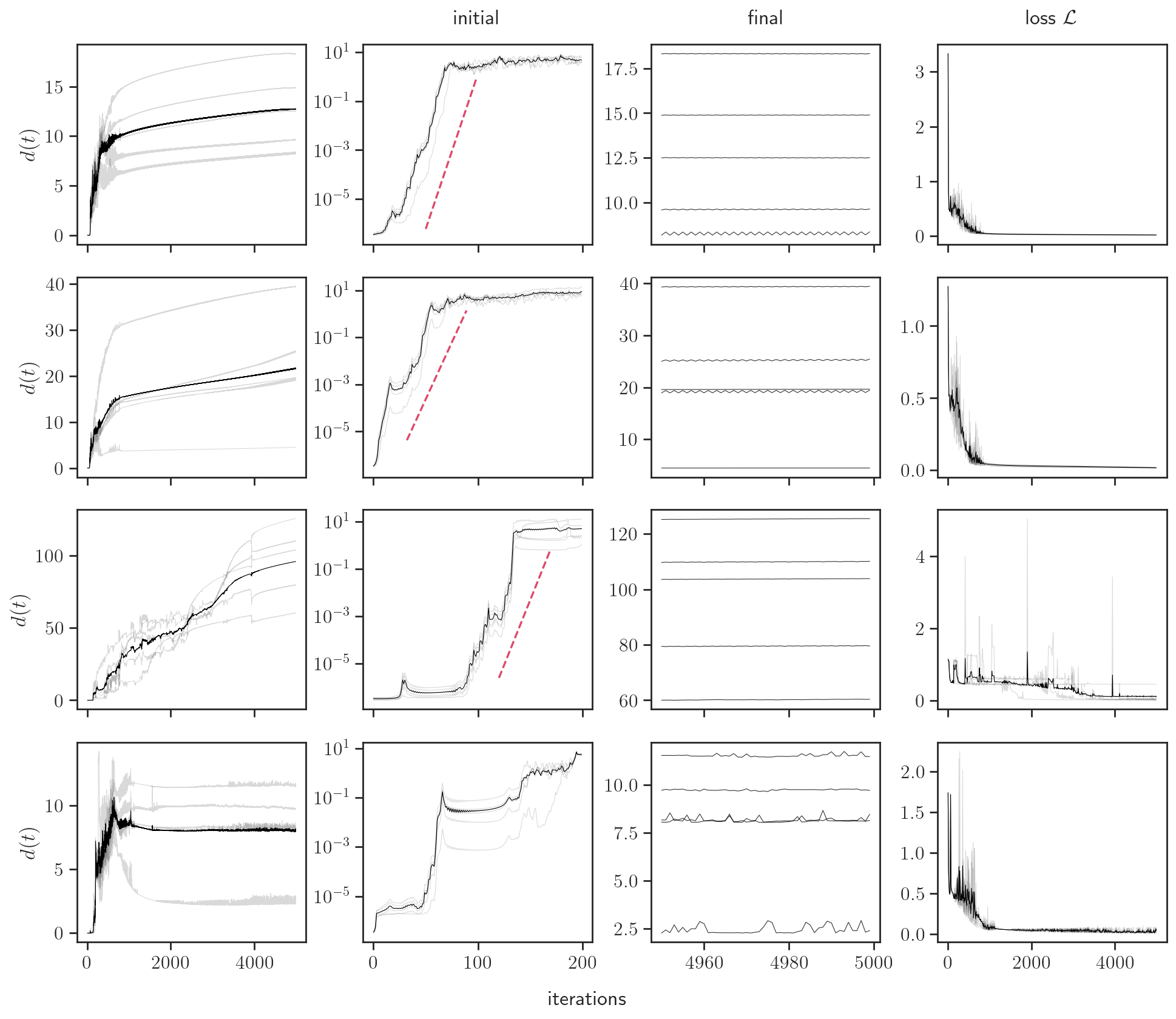}}
    \caption{Distance from perturbed to reference trajectories, and training loss in the Edge of Stability ($\eta=1$) regime, for four example initial conditions (each row corresponds to a different network initial condition). Around each initial condition, we build five perturbed networks, with $\epsilon=10^{-8}$. The first column from the left shows distances $d(t)$ for the full training period. The second column shows the region of exponential divergence for the first 200 iterations. Red dotted lines illustrate the slope of best fit for the exponential region. The third column shows the distances for the last 50 iterations. The final column shows the evolution of the loss $\mathcal{L}$.}
	\label{fig:eos}
\end{figure}

\section{The large learning rate regimes}\label{Sec:high_learning}

In the previous section, we explored the evolution of network trajectories when the learning rate is ``conventionally'' small ($\eta=0.01$), i.e., for which the loss function monotonically decreases towards its minimum by the action of the gradient descent scheme, see e.g. Fig. \ref{fig:ex}. Here we relax this assumption and consider larger values of the learning rate $\eta$, where such convergence is less well understood, and explore how both network trajectories and loss trajectories evolve and what type of dynamics are observed.

\subsection{$\eta=1$ (edge of stability): evidence of sensitive dependence on initial conditions}\label{ssec:eos}
Recent literature \cite{cohenGradientDescentNeural2021,agarwalAccelerationFractalLearning2021,kongStochasticityDeterministicGradient2020} points to the fact that gradient descent on non-convex loss functions does not necessarily become unstable when the learning rate is increased above the threshold predicted by the theory for convex optimization; astonishingly, within some region labelled the `edge of stability' the convergence of the neural network (i.e. learning) is faster than for the traditionally lower learning rate. 

\begin{figure}[htb!]
	\centering
	\includegraphics[width=\textwidth]{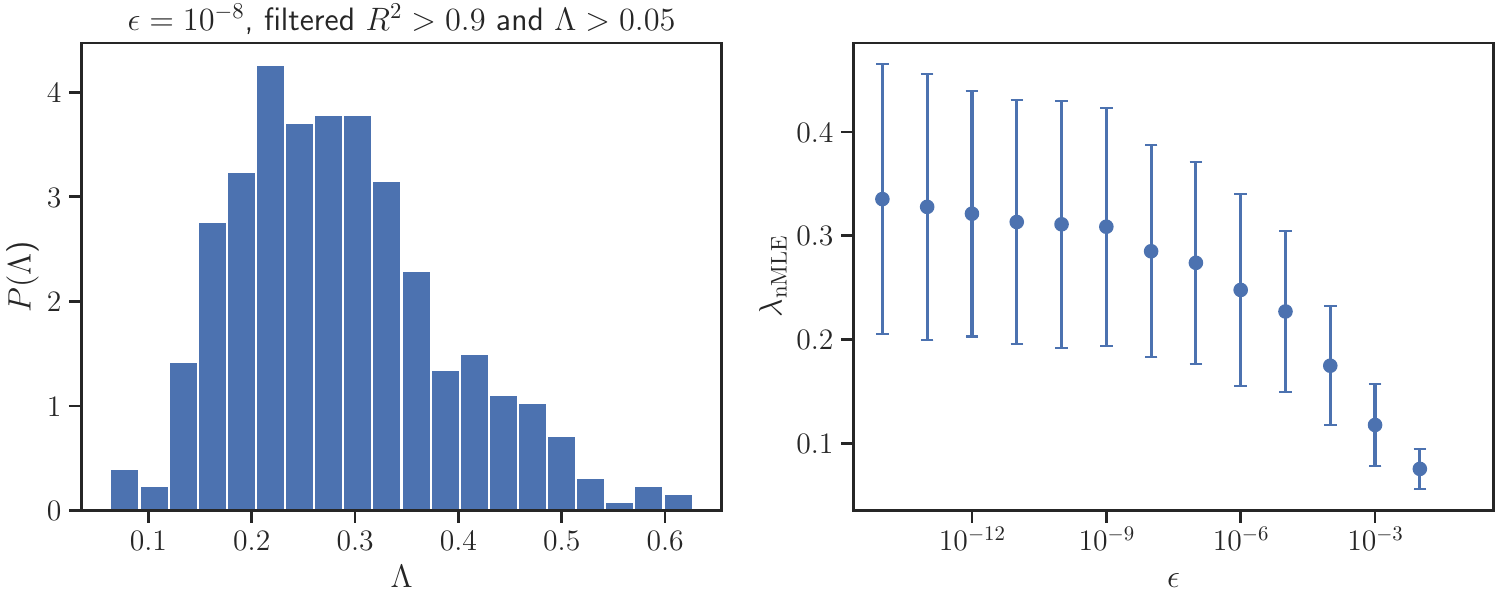}
	\caption{Lyapunov exponents for trajectories in the Edge of Stability ($\eta=1$) regime. (Left panel) Distribution of finite Lyapunov exponents $P(\Lambda)$, where each $\Lambda$ is estimated from Eq. \ref{eq:Lambda} for an $\epsilon$-ball of radius $\epsilon=10^{-8}$  centred at a network initial condition with 5 perturbed networks. $P(\Lambda)$ reconstructs the distribution for 500 different network initial conditions (only exponential fits with $R^2>0.9$ and value greater than $0.05$ have been used in order to exclude cases where no exponential divergence can be observed). (Right panel) Mean and standard deviation of $P(\Lambda)$, providing the network Maximum Lyapunov Exponent $\lambda_{\text{nMLE}}$ and its fluctuations, respectively, for $\epsilon\in[10^{-14},10^{-2}]$. $\lambda_{\text{nMLE}}$ converges to a stable value $\lambda_{\text{nMLE}}\approx 0.33$ as $\epsilon$ decreases. }
	\label{fig:eos_mle}
\end{figure}

 More concretely \cite{cohenGradientDescentNeural2021}, in this regime the maximum eigenvalue of the loss function Hessian (so called \textit{sharpness}) increases until it reaches the theoretical bound of divergence for Gradient Descent, yet the training itself does not diverge, and this occurs consistently across many tasks and architectures. In this regime, the loss overall tends to decrease but non-monotonically in the short-term. 

\medskip
\noindent Accordingly, we now fix a substantially higher learning rate\footnote{Note that $\eta=1$ is actually larger than the learning rate for which Cohen et al. \cite{cohenGradientDescentNeural2021} observe the Edge of Stability in their tasks. The authors note that for shallow networks and easy tasks, the sharpness increases to a lesser degree. Since our network is shallow and our task easy, it is reasonable to assume that we need larger values of $\eta$ to reach the sharpness necessary for the Edge of Stability. Additionally, they show that for cross-entropy loss (rather than e.g. MSE loss), the sharpness drops towards the end of training.} $\eta=1$ and replicate the analysis performed in Fig. \ref{fig:ex}. Results are plotted in Fig. \ref{fig:eos}, for four different network initial conditions and a perturbation radius $\epsilon=10^{-8}$. We can  see (fourth column) that while the loss eventually reaches a minimum close to zero, its transient behavior is clearly non-monotonic. At the same time, we can observe (first column) that the distance between initially close network trajectories typically show strong divergences (with an order of magnitude significantly larger than for $\eta=0.01$). Zooming in (second column), we can see that the initial divergence between network trajectories displays in many cases an exponential phase, whereas asymptotically (third column) distances are often oscillatory with a small period, sometimes resembling (stable) limit cycles. 

\medskip
\noindent We now pay a bit more attention to the presence of an exponentially expanding phase depicted in the second column of Fig. \ref{fig:eos}, which might be indicative to the presence of sensitivity to initial conditions in network space. One can quantify this effect by estimating the finite network Lyapunov exponent distribution $P(\Lambda)$ \cite{caligiuriLyapunovExponentsTemporal2023} {(the network version of finite Lyapunov exponents \cite{aurell1996growth}, following the procedure depicted \cite{caligiuriLyapunovExponentsTemporal2023} and briefly summarised in Section \ref{Sec:Preli})}.  In Fig. \ref{fig:eos_mle} (left) we show $P(\Lambda)$ reconstructed from 500 different network initial conditions, where (i) around each network initial condition we consider an $\epsilon$-ball of radius $\epsilon=10^{-8}$ and track the evolution of 5 perturbations within that radius, (ii) we compute $\Lambda$ via Eq. \ref{eq:Lambda}, where (iii) $\tau$ is automatically found as the window yielding the best exponential fit. We only keep those cases where the exponential fit has a $R^2 > 0.9$, and also discard cases where the resulting $\Lambda <0.05$ (around 90\% of the initial conditions were kept after this filtering was performed).
Observe that the distribution is unimodal, its mean is therefore a good proxy for the network MLE. In the right panel of Fig. \ref{fig:eos_mle}, we plot the resulting $\lambda_{\text{nMLE}}$, as a function of the perturbation radius $\epsilon$. The exponent stabilises to a positive value $\lambda_{\text{nMLE}}\approx 0.33$ as $\epsilon$ decreases, indeed suggesting the onset of sensitive dependence of initial conditions along the training process for $\eta=1$, an interesting result that clearly deserves further investigation.

\begin{figure}[htb!]
	\centering
        \includegraphics[width=0.9\textwidth]{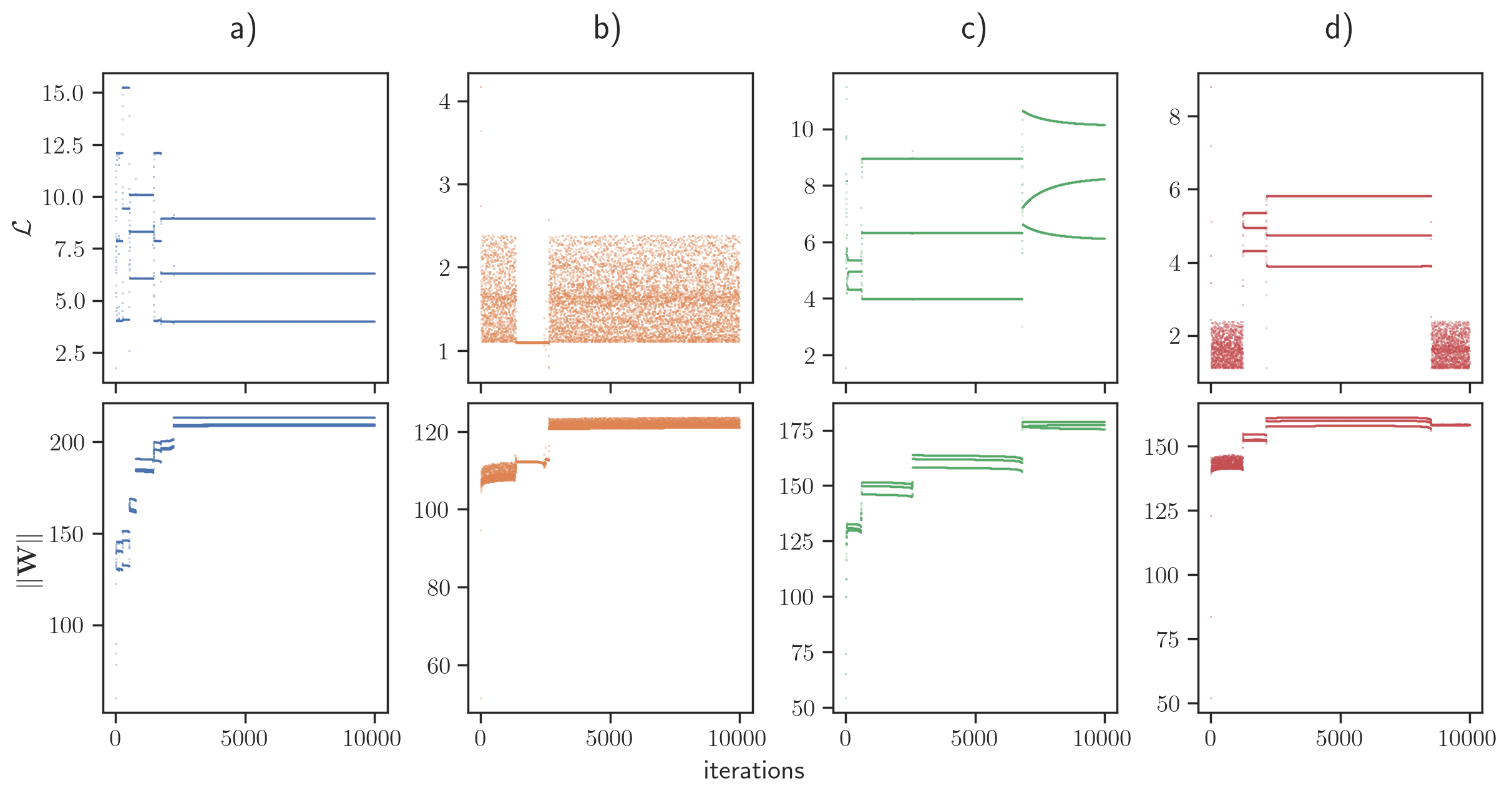}
	\caption{Training trajectories in the very large $\eta=5$ regime. Each column a)-d) represents the trajectory starting from an independent initial condition, where in total four were picked to illustrate the range of dynamical behaviors.  Shown are the training loss $\mathcal{L}$ (top row) and weight norms $\|\bf{W}\|$ (bottom row) for each trajectory.}
	\label{fig:chaotic}
\end{figure}

\subsection{$\eta=5$: rich taxonomy of dynamical behavior and hints of intermittency}
\label{ssec:chaos}

If we increase the learning rate well beyond the point that is conventionally considered stable, we still see no numerical divergence in the loss, but the dynamics --both at the level of the loss function and the network dynamics-- once again change dramatically. For illustration, Fig. \ref{fig:chaotic} shows the evolution of the loss $\mathcal{L}$ and the weight norm $\|\bf{W}\|$ for different independent initial conditions and a learning rate $\eta=5$. Generally, we can observe that the loss is very different from the other regimes studied so far. 
Its magnitude is very large compared to other regimes we have studied, to the point where it is difficult to argue that the network is indeed learning, except for occasions where the loss manages to settle to a small value. Even then, it is unclear whether the loss stays small for long times, since we sometimes observe jumps to larger-loss regions.\\
Interestingly, the time series of the loss for individual trajectories switches between a (period 3) quasi-periodic behavior\footnote{The time series does not strictly bounce between three fixed values, but instead three fixed regions. Within a region, the trajectory takes values that in isolation appear as a monotonically decaying series. The distance between region boundaries is usually $O(1)$ or larger, but the size of the regions is much smaller, often approximately $O(10^{-6})$. Thus, to the ``naked eye'' the trajectories appear periodic, e.g. on the plots in Fig. \ref{fig:chaotic}.} and a random-like phase. Such tendency for the trajectory to alternate between a quasi-periodic phase and a random-like phase is reminiscent of deterministic intermittency \cite{schuster2006deterministic, nunez2013horizontal}, a classical phenomenon describing the alternation of laminar phases intertwined with chaotic bursts. We thus look more closely into these random-like phases. 

\begin{figure}[]
    \centering
    \makebox[\textwidth]
    {\includegraphics[width=0.95\textwidth]{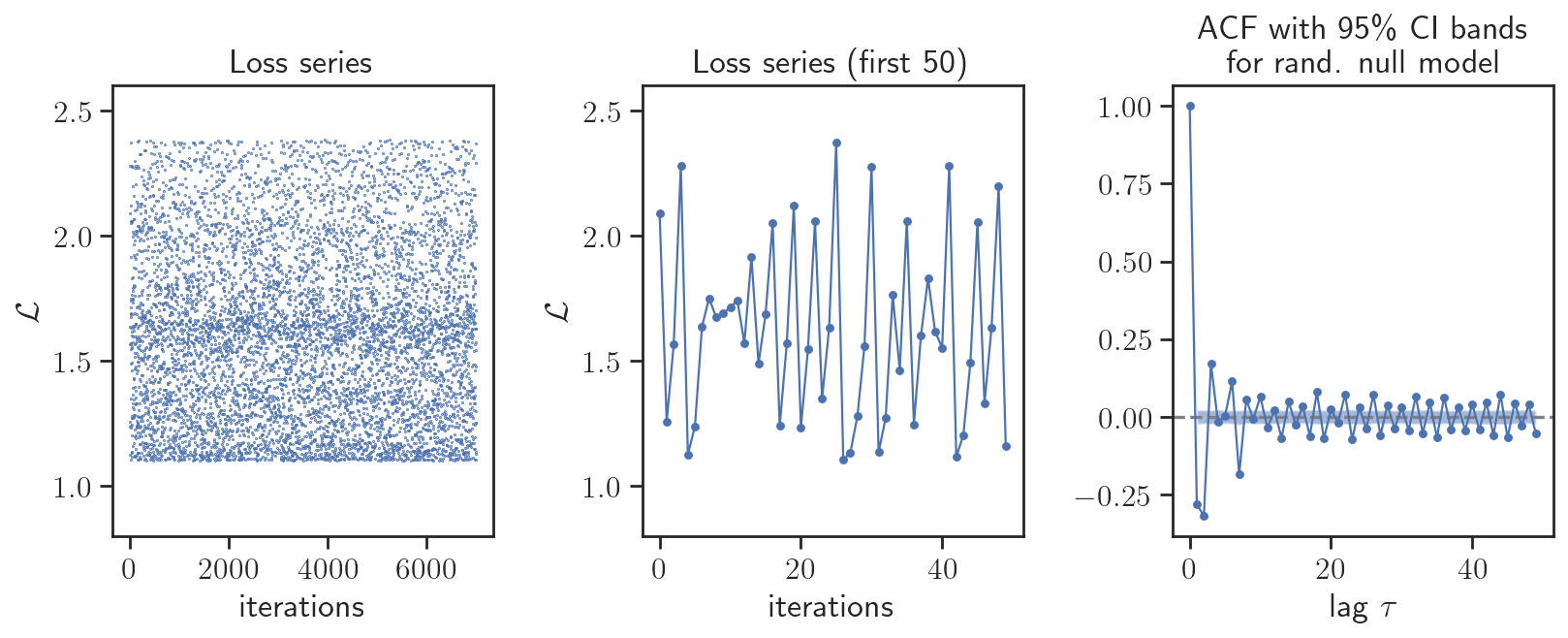}}%
    \caption{Closer look at a representative sequence of chaotic-like loss of a single trajectory (extracted from trajectory shown in column b) in Fig. \ref{fig:chaotic}) in the very large $\eta=5$ regime. (Left) Time series of loss during the chaotic-like regime. (Center) Zoom in on 50 iterations of the loss series. (Right) ACF up to lag $\tau=50$ of the loss series. The shaded area represents the bounds of the 95\% confidence interval for a randomized null model, i.e. the ACF computed for 1000 realizations of the shuffled series.}
    \label{fig:chaotic_acf}
\end{figure}
\begin{figure}[]
    \centering
    \makebox[\textwidth]
    {\includegraphics[width=0.8\textwidth]{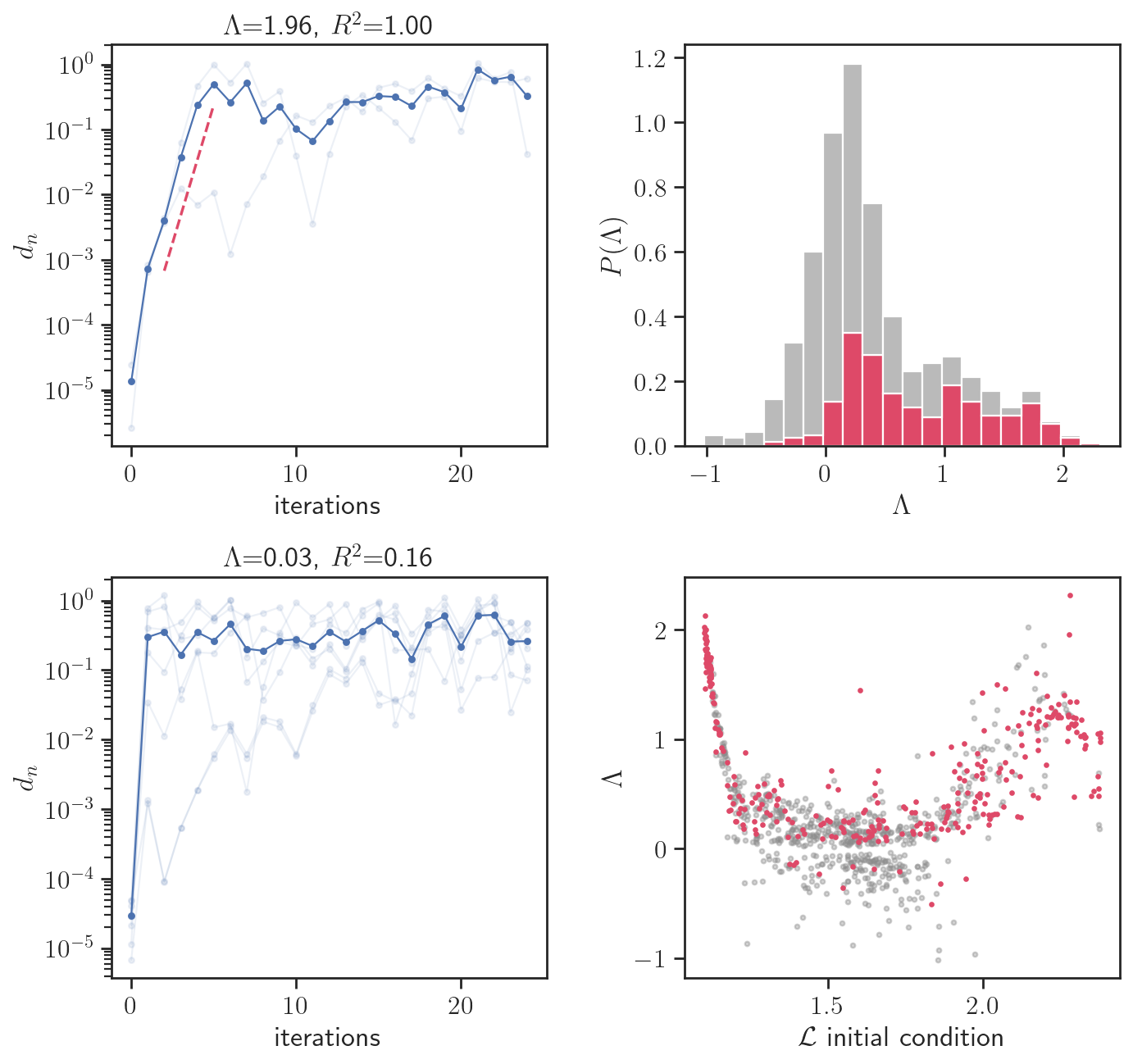}}%
    \caption{Analysis of local expansion rates for the chaotic-like loss of a single trajectory, shown in Figure \ref{fig:chaotic_acf}, in the $\eta=5$ regime. (Left panels) Divergence of initially nearby orbits for two example initial conditions. Light blue corresponds to distance $d_n$ between the initial condition and a single perturbation, while dark blue is the average distance over all perturbations for the given initial condition. The top panel shows an initial condition with a statistically significant exponential slope (best fit illustrated in red dashed line). The bottom panel shows an initial condition for which a slope of zero (i.e. no exponential divergence) cannot be rejected. (Right panels) Distribution of finite Lyapunov exponents $\Lambda$, for 1000 different initial conditions of the loss time series (top) and scatter plot of $\Lambda$ as a function of the initial condition of the loss $\mathcal{L}$ (bottom). Gray corresponds to initial conditions for which random evolution cannot be rejected (p-value $> 0.05$), while red corresponds to those for which a period of exponential divergence is statistically significant (p-value $< 0.05$).}
    \label{fig:chaotic_lambda}
\end{figure}
\begin{figure}[]
	\centering
 	\includegraphics[width=0.8\textwidth]{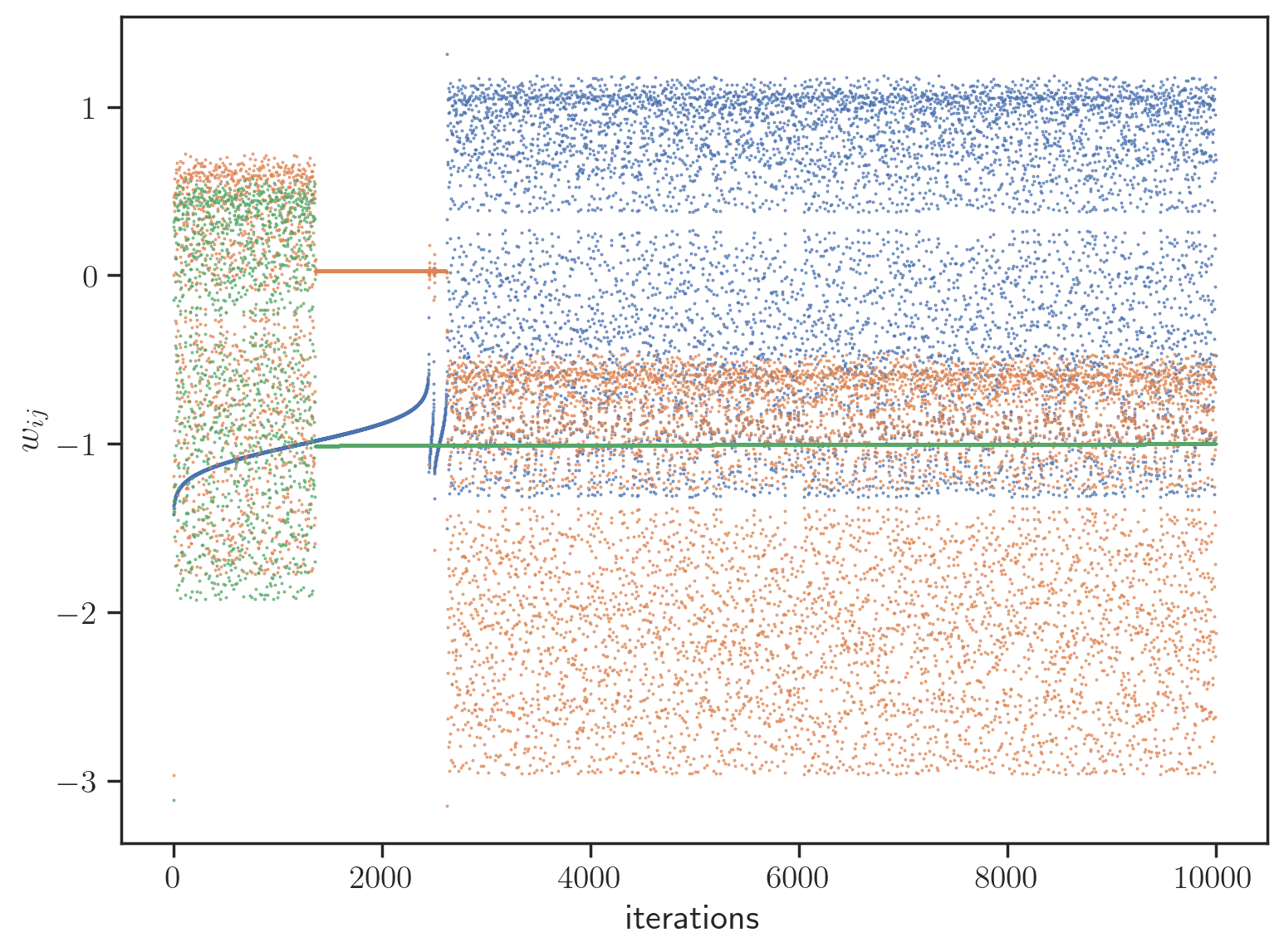}
	\caption{Trajectories for three individual weights that exhibit intermittent-like behavior, from a single initial condition (corresponding to column b) in Fig. \ref{fig:chaotic}). Different colors correspond to different weights $w_{ij}$.}
	\label{fig:chaotic_w}
\end{figure}

In Fig. \ref{fig:chaotic_acf} we depict the evolution of the scalar loss function, along with an estimation of its autocorrelation function, for a typical time series within one of these random-like intermissions. It is interesting to see that, while the time series is highly irregular and no obvious pattern emerges at the naked eye, the autocorrelation function detects statistically significant, periodic-like autocorrelation, suggesting that the loss time series might be performing an irregular evolution but alternating between two separate regions of the loss. This image is for instance reminiscent of the evolution of a chaotic orbit in a two-band chaotic attractor \cite{nunez2012detecting}. Subsequently, in Fig. \ref{fig:chaotic_lambda} we perform a Kantz-based \cite{kantz1994robust} approach to compute the (finite) Lyapunov exponent directly from the loss time series. Results indicate that for some initial conditions, there is evidence of sensitive dependence on initial conditions, whereas for many other initial conditions, such evidence is not statistically significant. All this points to the fact that the complex, intermittent-like evolution of the loss function cannot be simply accommodated to a one dimensional chaotic intermittent process. In hindsight, this is not surprising as the projection of the network dynamics into the loss function dynamics is quite severe: whereas the loss time series is a one-dimensional scalar one, the actual underlying system is high-dimensional and thus we expect a spectrum of Lyapunov exponents govern the long-run behavior. 
Finally, we observed that a qualitatively similar phenomenology is observed for the evolution of individual network weights $w_{ij}$ (Fig. \ref{fig:chaotic_w}). At the network level (Fig. \ref{fig:chaotic}), the observed intermittent-like behavior seems to be caused by a few weights acting in an intermittent-like fashion (which we have picked out for the plot in Fig. \ref{fig:chaotic_w}), while the rest of the weights remain constant throughout training. Interestingly, the onset or end of chaotic-like behavior seems to coincide for different weights. This rich phenomenology deserves further investigation, alongside with an investigation of the transition between the $\eta=1$ and the larger values explored here.

\section{Discussion and outlook}\label{Sec:discussion}

In this work we have illustrated how the process of training a neural network can be interpreted as a graph dynamical system yielding network trajectories, and how classical concepts from dynamical systems such as dynamical or orbital stability can be leveraged to gain some understanding of this training process \cite{lacasaCorrelationsNetworkTrajectories2022a, caligiuriLyapunovExponentsTemporal2023, ziyin2023probabilistic}.
For illustration, we considered a simple (toy) classification task, and trained a shallow neural network via gradient descent optimization. We analysed both the loss function time series and the actual neural network trajectories, and examined how small network perturbations propagate throughout the action of the training process to gain insights on dynamical and orbital stability of the graph dynamics. Our analysis allows us to distinguish clearly between two regimes, the so-called low learning rate regime ($\eta=0.01$) where gradient descent schemes are typically producing monotonically decreasing loss functions towards a minimum, and the large learning rate regime ($\eta \ge 1$) where such convergence is not necessarily guaranteed, and more complex dynamics at the level of the loss function can develop. Overall our results challenge naive expectations from low-dimensional dynamical systems and optimization theory.

\medskip
\noindent
In the low learning rate regime, despite the fact that the loss monotonically decreases towards a minimum, initially closeby network trajectories perform non-trivial evolution in graph space, marked by an alternation of divergence and convergence, eventually reaching a phase of slow yet monotonic divergence. Such behavior was found to be qualitatively similar regardless how close the network trajectories were initially but heavily dependent on the position of the initial condition within the whole graph space and results were put in the context of a lack of orbital stability. Similarly, we examined the evolution of closeby network trajectories in a post-learning process, i.e. where the loss had already approached a minimum, mimicking the dynamical stability analysis of dynamical systems close to a stationary point. We found hints of dynamical stability but overall results were pointing to the existence of plenty of irrelevant dimensions in graph space, i.e. the loss function minima being  more of a stationary manifold in graph space. 
The absence of (exponentially fast) convergence of perturbations of network stationary points deserves further investigation. Our conjecture that the marginal stability we observe is caused by the presence of flat dimensions (where the loss gradient vanishes) seems plausible in light of research presented earlier on the high-dimensional nature of low-loss basins \cite{fortLargeScaleStructure2019}. {In fact, there is evidence from both analytical results (in a reduced setting) and numerical experiments (for the Fashion-MNIST dataset) for the existence of wide and flat loss minima which, although rare, can be reached by many simple learning algorithms, especially when a cross-entropy loss function is employed \cite{baldassiShapingLearningLandscape2020}.} Intuitions developed from low-dimensional landscapes do not seem to hold for higher dimensions.\\
We stress that the phenomenology in the low learning rate regime is quite dependent on the initial condition, pointing to a severe loss of ergodicity, as it is usually the case for optimization problems in non-convex loss function landscapes. An interesting question for future work is the effect of including a regularization term in the loss function, which would essentially add a preferred direction for optimization in flatter regions of the loss landscape and thus we suspect would lead to more stable trajectories.

\medskip
\noindent 
{When the learning rate is large but the loss function still converges ($\eta=1$), we found hints of complex behavior both in the loss function time series and the network trajectories, including non-monotonic loss dynamics and hints of sensitive dependence of initial conditions for the network dynamics. Further research is necessary to elucidate whether this phenomenon is universal, but preliminary results in this direction (shown in Appendix \ref{app:mnist}) suggest that for the more complex MNIST dataset \cite{lecunGradientbasedLearningApplied1998}, network trajectories exhibit similar behavior and a region of optimal exploration-exploitation tradeoff with sensitive dependence on initial conditions is again identifiable. At this point, it is stimulating to mention the so-called edge of chaos paradigm, where dynamics poised near a critical point that separates an ordered and a disordered phase might evidence some degree of optimality in information processing capabilities \cite{carroll2020reservoir}. This classical hypothesis was introduced by Langton in the context of cellular automata \cite{langton1990computation}, and has been recently explored in the context of information processing \cite{boedecker2012information, carroll2020reservoir, vettelschoss2021information}. A very similar hypothesis is that living systems exhibit self-organized criticality \cite{bak1988self, watkins201625, hidalgo2014information, munoz2018colloquium}, with the brain being an archetypical example \cite{chialvo2010emergent, moretti2013griffiths, morales2023quasiuniversal}. Connecting the apparent criticality of brain dynamics with the information processing advantages of artificial systems and neural networks at the edge of chaos \cite{morales2023unveiling,carroll2020reservoir,morales2021optimal} has invigorated this interdisciplinary research line even further. It is thus suggestive to relate this phenomenology to our findings in the so-called edge of stability: a region where the loss function is still converging to a minimum (i.e., the ANN learns) albeit in a non-monotonic and faster way. The fact that in this region we find evidence of sensitive dependence on initial conditions (with positive maximum Lyapunov exponent) suggests that the search algorithm has switched from being a pure exploitation one for low learning rates to a balanced exploitation-exploration one at higher learning rate: a possible optimal strategy given the fact that the loss function indeed converges faster.}

\medskip
\noindent Finally, in the (very) large learning rate, we have observed and alternation of quasi-periodic and chaotic-like evolution of both the loss and the network itself (pointing to the presence of chaotic intermittency) for even larger learning rates. Further research is needed to better understand the dynamical nature of these regimes, their possible relation to classical paradigms of complex behavior such as the intermittency routes to chaos, and how these could be leveraged to develop deterministic gradient-based training strategies at extremely large learning rates \cite{geipingStochasticTrainingNot2022, kongStochasticityDeterministicGradient2020}.

\medskip
To conclude, this work provides an illustration as to how concepts and tools from dynamical systems, time series analysis and temporal networks can be combined to gain understanding of the training process of a neural network. The specific classification task and network architecture under consideration were chosen for illustration, rather than specific interest. In this sense, more realistic scenarios (both for tasks and network architectures) should be explored. Further work is also needed to understand whether the results presented here generalize well across tasks and architectures, or e.g. if specific architectures display different types of dynamical stability.
Ultimately, our exploratory findings aims to stimulate research and exchange of ideas between the above-mentioned fields.

\backmatter

\bmhead{Acknowledgments} Authors acknowledge helpful feedback from Manuel Matias and Massimiliano Zanin and IFISC MSc in Physics of Complex Systems. We acknowledge funding from the Spanish Research Agency MCIN/AEI/10.13039/501100011033 via projects DYNDEEP (EUR2021-122007), MISLAND (PID2020-114324GB-C22), INFOLANET (PID2022-139409NB-I00) and the María de Maeztu project CEX2021-001164-M.

\medskip
\noindent{\bf Data and code.} Upon publication, data and code will be available at \url{https://github.com/GitKalo/ann_dynamics}

\appendix


\section{Performance on Iris dataset}
\label{app:iris}

\begin{figure}[h]
    \centering
    \includegraphics[width=0.9\textwidth]{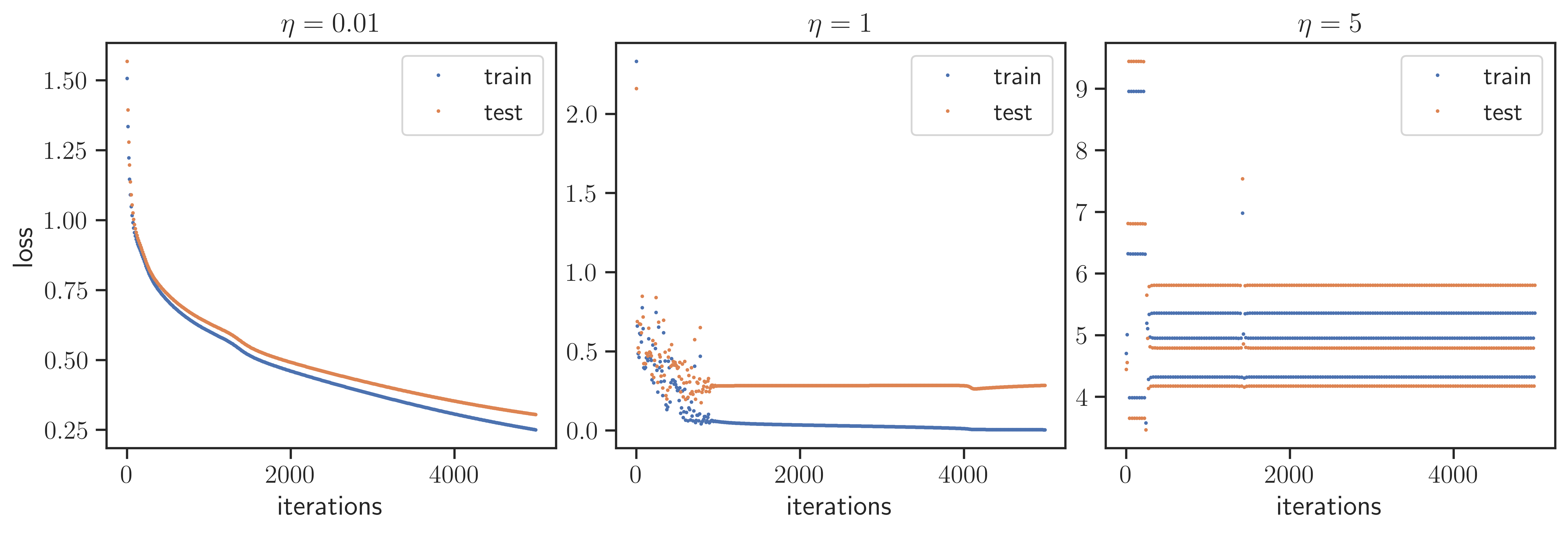}
    \caption{{Example training and testing loss of network on the Iris dataset for different values of the learning rate $\eta$. Results are for a single initial condition in each panel.}}
    \label{fig:iris_loss}
\end{figure}

\begin{figure}[h]
    \centering
    \includegraphics[width=0.9\textwidth]{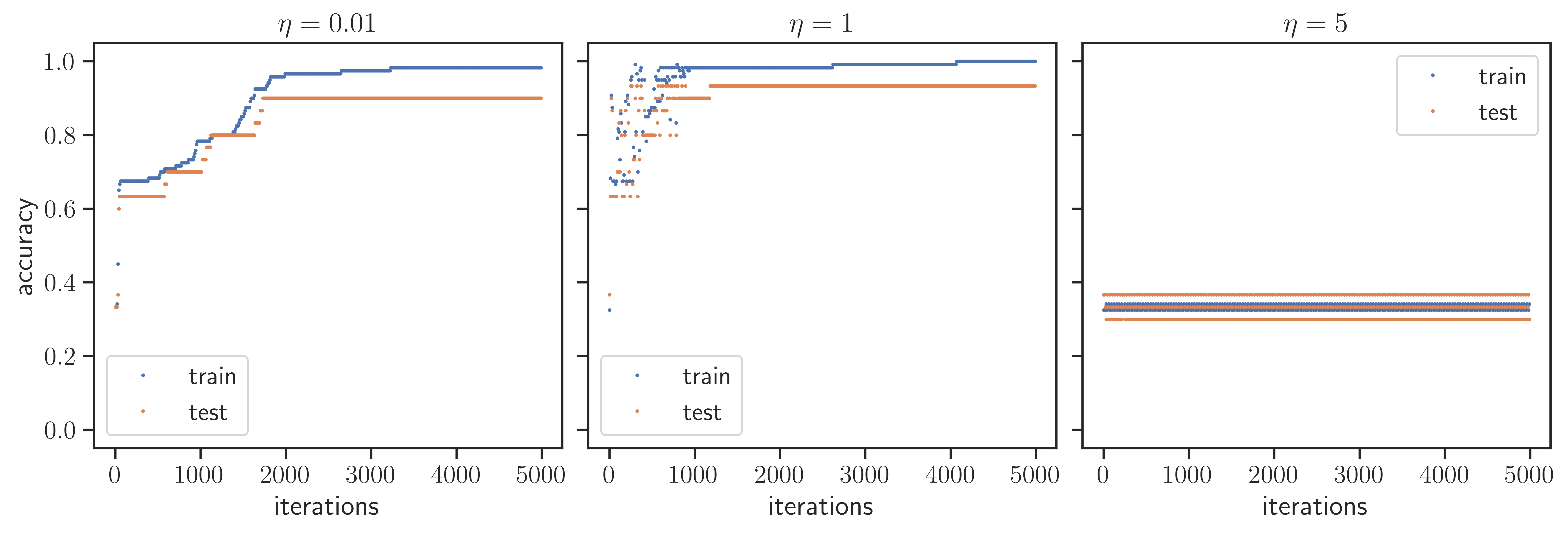}
    \caption{{Example training and testing accuracy of network on the Iris dataset for different values of the learning rate $\eta$. Results are for a single initial condition in each panel.}}
    \label{fig:iris_acc}
\end{figure}

Here we show sample performance metrics for the network architecture used in this paper, in the different regimes of the learning rate $\eta$ explored previously. The full dataset of 150 samples was split into a fixed training dataset of 120 datapoints and testing dataset of 30 datapoints. We show results for the network CE loss (Fig. \ref{fig:iris_loss}) and accuracy (Fig. \ref{fig:iris_acc}) on both the training and testing dataset. The position of the panels corresponds to the same initial conditions in the two figures.

In general, for the low ($\eta=0.01$) and medium ($\eta=1$) regimes, the network managed to learn the classification task effectively, while in the large ($\eta=5$) regime, there is little learning happening (the network is essentially a random classifier). The loss shows a generalization gap that seems larger in the edge of stability ($\eta=1$) regime, although this is not reflected in a lower accuracy on the test set.

\section{Preliminary analysis on MNIST dataset}
\label{app:mnist}

In an effort to move towards a more universal characterization of ANN training dynamics, in this section we expose some results of applying the methodology described in this paper on the more complex MNIST dataset. We explore the different regimes defined by the learning rate $\eta$, and the networks' performance, finding important similarities and interesting differences with the results presented before for the Iris task.

The MNIST presents the challenge of classifying images of handwritten digits, and is a standard problem used to calibrate learning models. For the results presented below, we use an architecture consisting of a single hidden layer of $H=64$ units with a sigmoid activation function. This results in a network of $\approx50$K trainable weights. Kernel weights are initialized as $\hat{N}(0,1)$ random numbers and bias weights are initialized to zero. As before, we do not use regularization or dropout, and perform full-batch training epochs (i.e. deterministic GD) on a fixed subset of 5000 samples from the MNIST training dataset. With this setup, the task and architecture are substantially more complex than the case studied before for the Iris, while remaining computationally tractable (recall that our analysis requires running hundreds of training sessions with potentially thousands of epochs each).

\begin{figure}[h]
    \centering
    \includegraphics[width=0.9\textwidth]{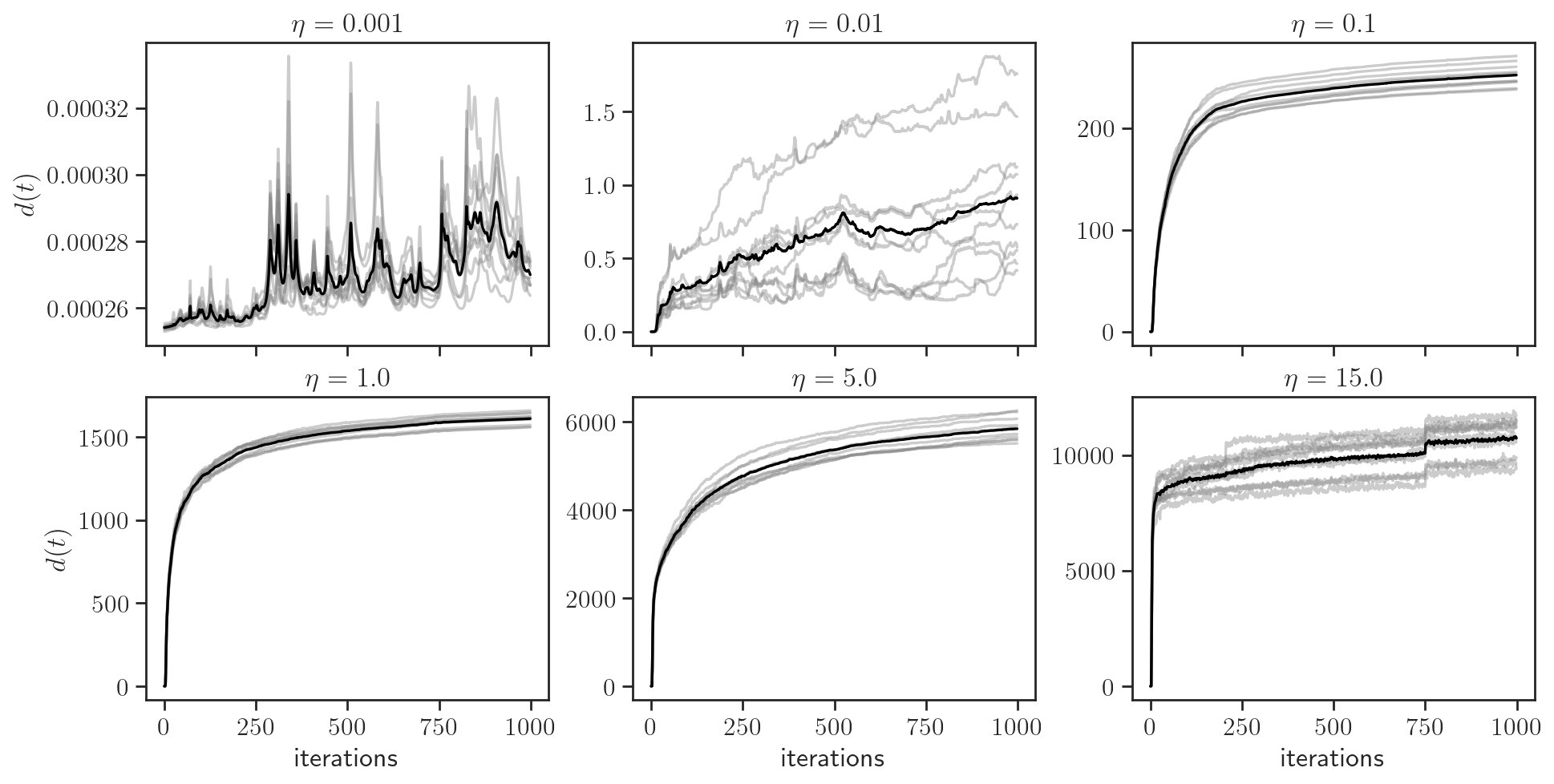}
    \caption{{Distances between reference and perturbed trajectories for different values of the learning rate $\eta$. Each panel shows results for a single initial condition, with distances from 20 individual perturbations (gray lines) and their average distance (black line) over 1000 epochs. Perturbations are taken with a perturbation radius $\epsilon=10^{-8}$.}}
    \label{fig:mnist_dists}
\end{figure}

In Figure \ref{fig:mnist_dists} we show the evolution of the distances between a reference trajectory and a ball of 20 perturbations taken at the initial condition. For small learning rates, we see irregular, non-monotonic behavior of the distances, with a general upward trend. For intermediate and larger learning rates, we see a quick increase in the distance, followed by a slow saturation to a value dependent on the learning rate. All of these results are consistent with those found for the Iris dataset (see for example Figures \ref{fig:ex} and \ref{fig:eos}), with the small exception that here the increase in distances seems less erratic for intermediate learning rates.

\begin{figure}[h]
    \centering
    \includegraphics[width=0.8\textwidth]{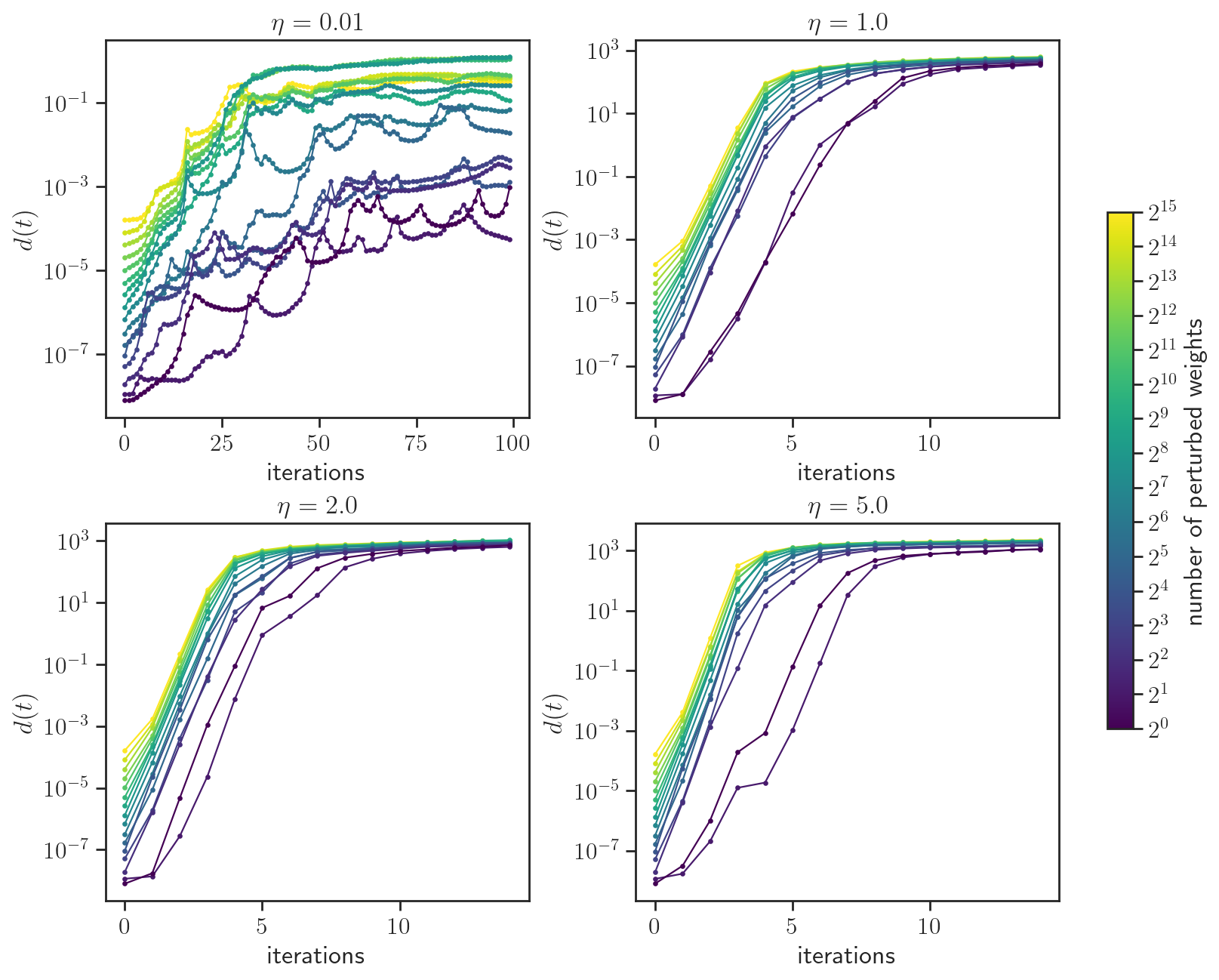}
    \caption{{Distances between reference and perturbed trajectories for different values of the learning rate $\eta$ in the $\epsilon\to0$ limit. The limit is explored by perturbing a different number of weights at initialization. Each panel shows results for a single initial condition, with colors corresponding to the number of perturbed weights. Perturbations are taken with a radius $\epsilon=10^{-8}$.}}
    \label{fig:mnist_eps_lim}
\end{figure}

Most importantly, a behavior that is consistent across both tasks is a region of exponential divergence of distances near the start of training, i.e. an indication of sensitive dependence to initial conditions. This is shown clearly in Figure \ref{fig:mnist_eps_lim}, where we explore the evolution of the distances near the initial condition in the limit $\epsilon\to0$. Since numerical precision does not allow us to explore this limit by taking the perturbation radius $\epsilon$ arbitrarily small, we instead proceed by fixing $\epsilon=10^{-8}$ and perturbing a progressively smaller number of weights at initialization. The distances for intermediate values of the learning rate $\eta$ show a clear exponential increase that becomes more pronounced as the initial distance (number of perturbed weights) becomes smaller.

\begin{figure}[h]
    \centering
    \includegraphics[width=0.85\textwidth]{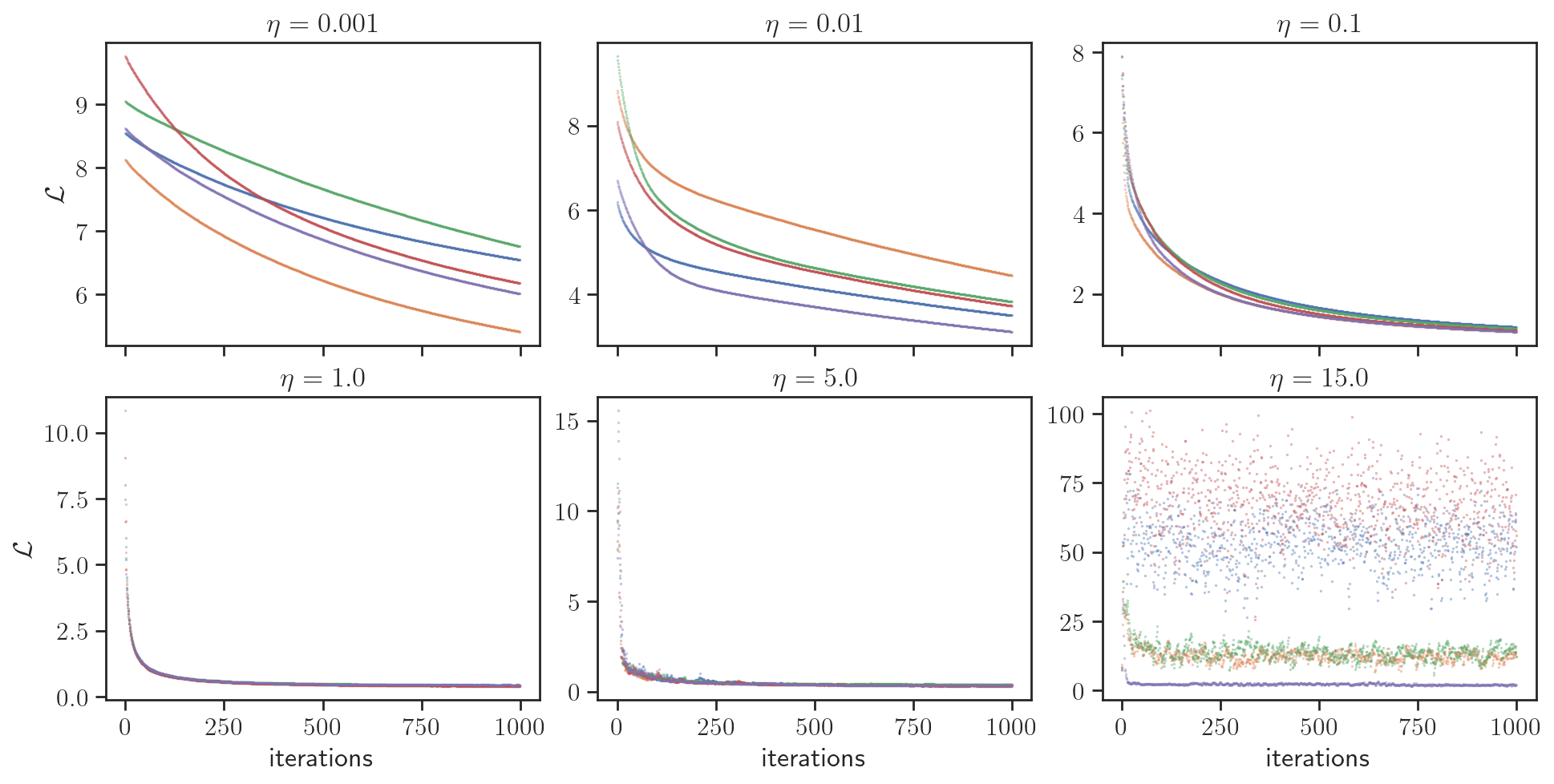}
    \caption{{Training loss on the MNIST dataset for different values of the learning rate $\eta$, with different colors corresponding to different, i.i.d initial conditions.}}
    \label{fig:mnist_loss}
\end{figure}

\begin{figure}[h]
    \centering
    \includegraphics[width=0.85\textwidth]{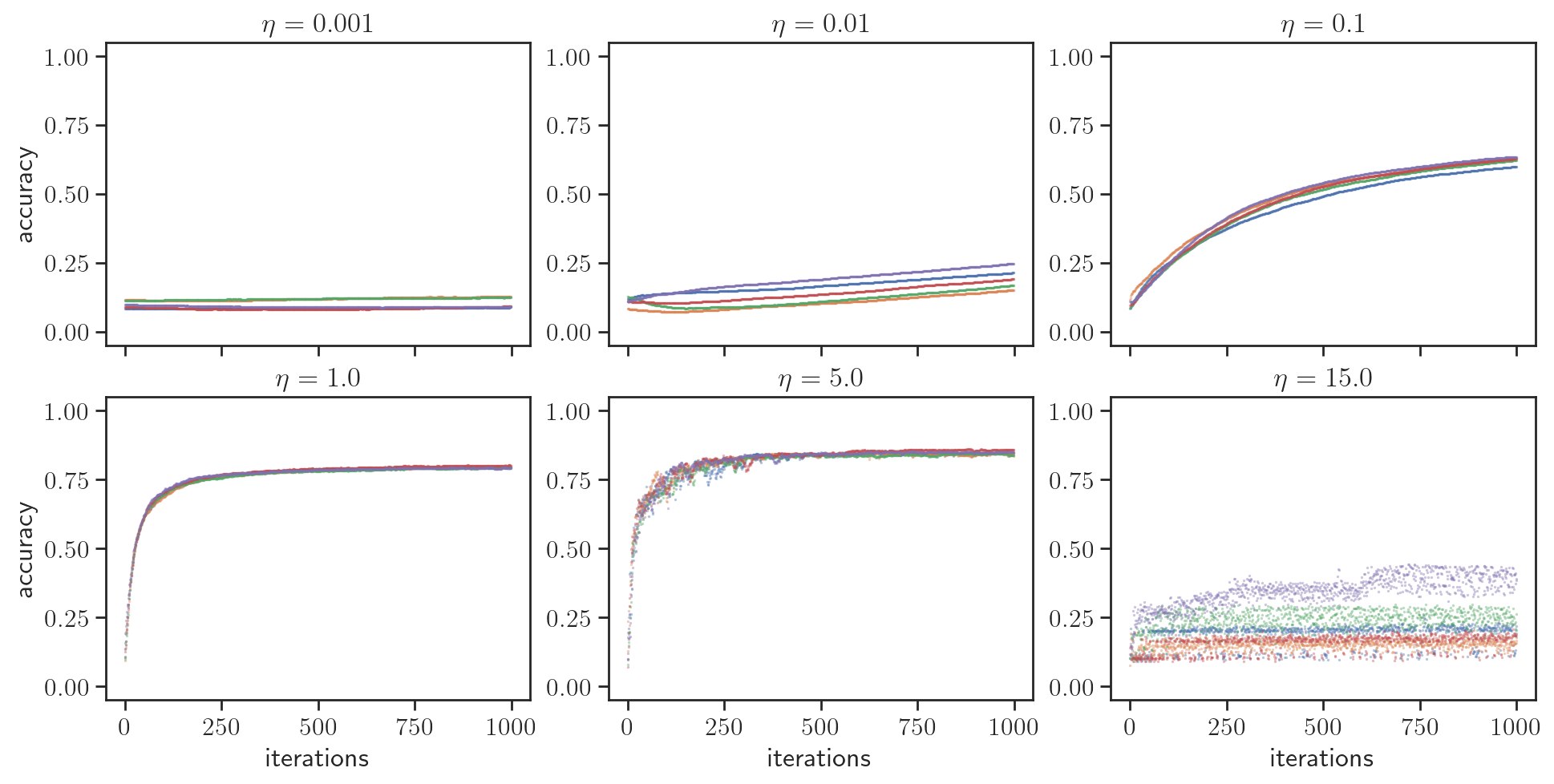}
    \caption{{Test set accuracy on the MNIST dataset for different values of the learning rate $\eta$, with different colors corresponding to different, i.i.d initial conditions.}}
    \label{fig:mnist_acc}
\end{figure}

In addition, by inspecting the training loss (Figure \ref{fig:mnist_loss}) and testing accuracy (Figure \ref{fig:mnist_acc}) we see that the regions where we find sensitive dependence on initial conditions ($\eta$ between 0.1 and 5, in this case) also correspond to faster and more optimal learning and (for $\eta=5$) non-monotonic decrease in the loss. This is consistent with the Edge of Stability region identified for the Iris dataset (see Section \ref{ssec:eos}), where training was faster but less regular. One important difference is that for the MNIST we see sensitive dependence on initial conditions even for learning rates that show slower, monotonic learning (e.g. $\eta=1$).

Note also the important differences between these results and those presented for the Iris. First of all, as we expected, the transitions between regular, irregular (edge of stability) and chaotic-like regimes seem to occur at different values of $\eta$, although of course a much more precise study is necessary to pinpoint the exact transition values. In addition, in the chaotic-like regime we do not see at first glance quasi-periodic or intermittent-like behavior, as we did for the Iris (see Section \ref{ssec:chaos}). Still pending is further analysis, similar to the one performed in \ref{Sec:high_learning}, to test for signatures of chaotic behavior in the loss and accuracy time series. Nevertheless, these results, although limited to a specific architecture and parameters, point towards the possible existence of universal behaviors for ANN training trajectories.

\color{black}

\clearpage


\bibliographystyle{sn-mathphys}
\bibliography{refs}

\end{document}